\documentclass[final,10pt,5p,times]{elsarticle}
\usepackage[svgnames,dvipsnames,table]{xcolor}
\usepackage{ifthen}
\usepackage{hyperref}
\usepackage{graphicx} % Required for inserting images
\usepackage[baseline]{euflag}
\usepackage{multirow}
\usepackage{array}
\usepackage{tabularx}
\usepackage{caption}
\usepackage{subcaption}
\usepackage{amsmath}
\usepackage{siunitx}
\usepackage[export]{adjustbox}
\usepackage{color, colortbl}
\usepackage{booktabs}
\usepackage{xspace}
\usepackage{ragged2e}
\usepackage{lineno}
\input{def_macros.tex}

\newcommand\Sm{\texttt{Simple} }
\newcommand\Smj{$\texttt{Simple}_\texttt{STD}$ }
\newcommand\Smw{$\texttt{Simple}_\texttt{WB}$ }
\newcommand\Ss{$\texttt{Season21ToSeason22Phone}$ }
\newcommand\Ssj{$\texttt{Season21ToSeason22Phone}_\texttt{STD}$ }
\newcommand\Ssw{$\texttt{Season21ToSeason22Phone}_\texttt{WB}$ }
\newcommand\Ssa{$\texttt{Season21ToSeason22All}_\texttt{WB}$ }
\newcommand\Ms{$\texttt{MixedSeason}$ }
\newcommand\Msj{$\texttt{MixedSeason}_\texttt{STD}$ }
\newcommand\Msw{$\texttt{MixedSeason}_\texttt{WB}$ }

\begin{document}

\begin{frontmatter}

\title{Can Robots ``Taste'' Grapes? Estimating SSC with Simple RGB Sensors}
\author[inst1, inst2]{Thomas A. Ciarfuglia}
\author[inst2]{Ionut M. Motoi}
\author[inst2]{Leonardo Saraceni}
\author[inst2]{Daniele Nardi}

\cortext[cor1]{Corresponding author: \href{mailto:thomas.ciarfuglia@uniroma5.it}{thomas.ciarfuglia@uniroma5.it}}
\affiliation[inst1]{organization={San Raffaele University of Rome
},%Department and Organization
            city={Rome},
            %postcode={00166}, 
            %state={RM},
            country={Italy}}
\affiliation[inst2]{organization={Sapienza University of Rome, Department of Computer, Control and Management Engineering
},%Department and Organization
            city={Rome},
            %postcode={00185}, 
            %state={RM},
            country={Italy}}

% \maketitle
\begin{abstract}
In table grape cultivation, harvesting depends on accurately assessing fruit quality. While some characteristics, like color, are visible, others, such as Soluble Solid Content (SSC), or sugar content measured in degrees Brix (°Brix), require specific tools. SSC is a key quality factor that correlates with ripeness, but lacks a direct causal relationship with color. Hyperspectral cameras can estimate SSC with high accuracy under controlled laboratory conditions, but their practicality in field environments is limited. This study investigates the potential of simple RGB sensors under uncontrolled lighting to estimate SSC and color, enabling cost-effective, robot-assisted harvesting. Over the 2021 and 2022 summer seasons, we collected grape images with corresponding SSC and color labels to evaluate algorithmic solutions for SSC estimation, specifically testing for cross-seasonal and cross-device robustness. 
We propose two approaches: a computationally efficient histogram-based method for resource-constrained robots and a Deep Neural Network (DNN) model for more complex applications. Our results demonstrate high performance, with the DNN model achieving a Mean Absolute Error (MAE) as low as $1.05$ °Brix on a challenging cross-device test set. The lightweight histogram-based method also proved effective, reaching an MAE of $1.46$ °Brix. These results are highly competitive with those from hyperspectral systems, which report errors in the $1.27$--$2.20$ °Brix range in similar field applications.
\end{abstract}

\end{frontmatter}

\section{Introduction} \label{sec:intro}

\begin{figure}[th!]
    \includegraphics[width=\columnwidth,left]{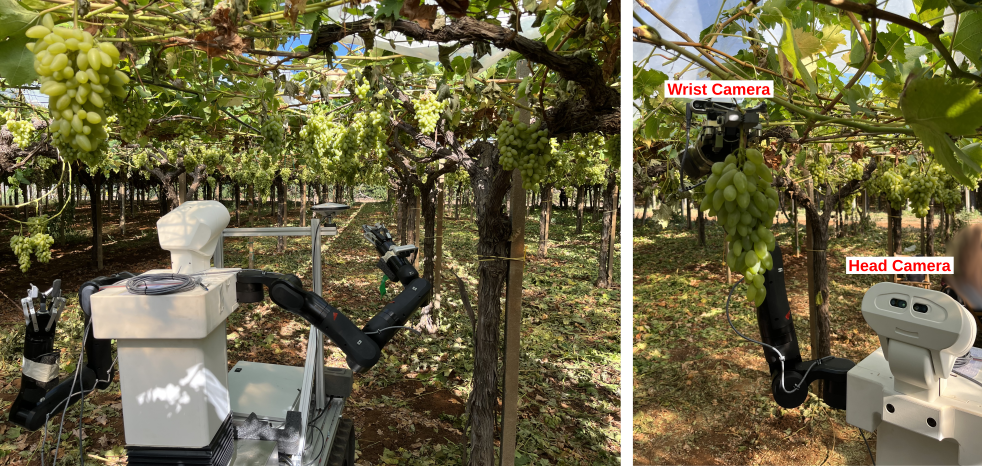}
    \caption{A robotic platform for grape harvesting developed within the CANOPIES project (\cite{canopies}), which provides the context for this research. Such robots require robust perception systems to assess fruit quality in the field using on-board sensors The end-effector and the head of the robot are equipped with simple, low-cost RGB cameras suitable for this task. This work focuses on developing and validating the core data-driven algorithms necessary for SSC and color estimation in these challenging robotic setups. }
    \label{fig:visual_abstract}
\end{figure}

The use of passive imaging sensors to measure agronomic properties of crops enables fast, non-destructive monitoring that, when integrated with machinery or robots, can be efficiently scaled to cover entire fields (Fig.~\ref{fig:visual_abstract}). RGB (Red, Green, Blue) cameras are commonly employed to monitor external conditions, such as detecting pests and diseases, while multi-spectral cameras are occasionally used to derive indices related to vegetation health. However, assessing internal crop properties presents a more complex challenge. Hyperspectral imagery (HSI) shows potential in this area, as its ability to capture specific wavelength responses can be correlated with the absorption patterns of certain chemical properties. Nevertheless, hyperspectral cameras are typically slow to acquire images, highly sensitive to illumination conditions, and usually unsuitable for use on operational agricultural machinery.

\newcommand{\High}{\textcolor{BrickRed}{High }}
\newcommand{\Low}{\textcolor{ForestGreen}{Low }}
\begin{table*}[h!]
    \small
    \centering
    \caption{Comparison of Hyperspectral (HSI) and RGB Imaging devices for in-field use.}
    \label{tab:hsi_vs_rgb}
    \begin{tabularx}{\textwidth}{@{} l X X @{}}
        \toprule
        \textbf{Feature} & \textbf{Hyperspectral Imaging (HSI)} & \textbf{Standard RGB Imaging} \\
        \midrule
        \textbf{Spectral Resolution}& 
        \High (hundreds of narrow bands, usually 5--10 nm) & 
        \Low (3 broad bands: Red, Green, Blue) \\
        \addlinespace 

        \textbf{Spatial Dimension}& 
        \textcolor{BrickRed}{Low} & 
        \textcolor{ForestGreen}{High} \\
        \addlinespace
        
        \textbf{Information Captured} & 
        Detailed spectral signatures, correlated with chemical composition & 
        Surface color, texture, and shape \\
        \addlinespace
        
        \textbf{Equipment Cost} & 
        €$10000$ - €$200000+$ & 
        €$10$ - €$1000+$ \\
        % from thousands to hundred thousands euros & 
        % from few euros to thousands euro \\
        \addlinespace
        
        \textbf{Operational Environment} & 
        Laboratory or highly controlled settings& 
        In-field or indoors alike \\
        \addlinespace

        \textbf{Acquisition Time} & 
        \High (seconds to minutes, depending on illumination) & 
        \Low (few milliseconds) \\
        \addlinespace
        
        \textbf{Computational Load} & 
        \High (due to the processing 3D hypercubes) & 
        \Low (suitable for real-time processing on embedded systems) \\
        \addlinespace
        
        \textbf{Portability \& Size} & 
        Often bulky and requires careful calibration & 
        Compact, and easily integrated into robotic platforms \\
        \bottomrule
    \end{tabularx}
\end{table*}

These competing aspects are at odds with the desirable tension to bring more robots to the farming fields, in order to both face the shortage of workforce, but also to manage the crops in a more precise way. 
In fact, many cultivations are labor-intensive, and for various economic and social reasons, it is increasingly difficult to find human workers to perform operations such as harvesting. Table grape cultivation is an example of such labor-intensive cultivation that is increasingly popular in the markets but faces a workforce shortage. This shortage would be mitigated by the introduction of robot workers that could, to some extent, assist human workers in the field. To be able to do so, the robotic platform should be able to assess the quality and ripeness of the fruit, such as color, Soluble Solid Content (SSC), size, presence of anomalies, using mainly passive sensors (active manipulation for perception would be orders of magnitude slower). The case of SSC estimation is of particular relevance as its estimation via passive sensors is usually done with Infrared (IR) refractometers (that require at least contact with the berries) or hyperspectral imaging (HSI) devices. Although solutions equipping robots with HSI cameras are becoming more prevalent, their high cost and operational complexity remain major barriers to widespread adoption compared to the use of other devices, such as RGB cameras (Table~\ref{tab:hsi_vs_rgb} offers a summary of the main differences between hyperspectral and RGB cameras). 
So, the question of designing a passive SSC estimation system using low-cost, reliable, RGB cameras, or similar devices is still open. 

This work addresses this question by validating its feasibility and proposing high-performance algorithmic solutions tailored for practical deployment in agricultural settings. In the following paragraphs, we provide a brief review of the relevant literature, highlighting key contributions to contextualize this study. For a broader overview of fruit ripeness estimation methods—particularly for grapes using hyperspectral and RGB imaging—we refer the reader to \citet{Palumbo2023Computer} and \citet{Vrochidou2021Machine}.

\begin{table*}[h!]
	\small
	\centering
	\caption{Comparison of different approaches for SSC estimation in Table Grapes using spectral equipment.}
	\label{tab:hsi_approaches}
	%\begin{adjustbox}{center}
	%\resizebox{\textwidth}{!}{
	    \setlength{\tabcolsep}{4pt} % Reduce space between columns (default is 6pt)
%	    \begin{tabularx}{\textwidth}{@{} l p{1.8cm} p{1.8cm} p{2.2cm} p{1.8cm} p{1.2cm} p{1.8cm} p{1.8cm} @{}}
	    \begin{tabularx}{\textwidth}{@{} l X X X X X X X @{}}
	    \toprule
	    \textbf{Name} & \textbf{Camera model} & \textbf{Camera type} & \textbf{Spectral Range} & \textbf{Spectral Res.} & \textbf{Spatial Res}. & \textbf{Cultivar} & \textbf{Illumination} \\
	    \midrule
	    %% row 1
	    \citet{gutierrez2019on-the-go} & Resonon Pika & Push broom & VIS-NIR (400-1000nm) & 2.1nm (300 bands) & 300 px & Tempranillo & Sunlight \\ % row 2
	    \citet{Kolapesa2023Estimation} & Contact probe spectrometer & Spectrometer (contact probe) & VNIR–SWIR (350–2500 nm) & Not specified & N/A (point measurement) & Chardonnay, Malagouzia, Sauvignon-Blanc, Syrah & Internal illumination \\
	    % row 3
	    \citet{Tsakiridis2023In-situ} & Cubert FireflEYE V185 & Snapshot & VIS-NIR (400–1000 nm) & 4-8nm (128 bands) & 50x50 & Chardonnay, Malagouzia, Sauvignon-Blanc, Syrah & Sunlight with learned calibration \\
	    \bottomrule
	    \end{tabularx}
	    %}
	%\end{adjustbox}   
\end{table*}

\subsection{Hyperspectral applications to agricultural robotics}
Robots have been used for the monitoring and quality estimation of different crops and are now finding some industrial applications in high-value crops (\eg wine grapes in \citet{Williams2023Modelling} and \citet{Polvara2024Bacchus}, strawberries in \citet{Xiong2019Development, Xiong2020Autonomous}, blueberries in \citet{Williams2024Barracuda}, among others). Some authors have explored specifically the quality estimation of fruits with hyperspectral sensors. In Table~\ref{tab:hsi_approaches} we summarize the HSI approaches applied to grapes internal quality estimation.

In \citet{gutierrez2019on-the-go} an HSI device is mounted on an Autonomous Ground Vehicle (AGV) to measure the Total Soluble Solids (TSS) --- a synonym of SSC --- of Tempranillo wine grapes. The AGV moves along the vineyard row acquiring spectral data in a push-broom fashion. To achieve color constancy the authors perform an indirect calibration for exposure, while illuminant calibration was performed every 5m with a white reference. Although the authors claim that this setup shows on-the-go HSI estimation feasibility, they also note the difficulty of maintaining the steady motion required by push-broom cameras using an AGV. In addition, white balancing stop-and-go operations are required every 5 meters to calibrate the camera. The impact of these choices on the accuracy of the estimation is not clear, and the authors report a Root Mean Square Error (RMSE) on °Brix estimate as low as $1.274$°. Finally, the method requires considerable post-processing time (5 hours), making it not well suited for robotic harvesting. Being able to use a simple RGB camera for the estimation would remove both the vibration and calibration issues.

A similar approach, but with a trailing cart instead of an AGV, was proposed by \citet{Williams2017Method}. The cart mounts two HSI devices (VNIR: 400-896nm, SWIR: 896-2506nm) to monitor raspberries. Again, the push broom image is obtained by moving a vehicle and the whole setup requires  an overarching arm with a black background screen behind the plants and  from which a white reference is hung. The aim of the study was to just demonstrate the setup and no physical characteristics have been estimated using the hyperspectral data.

A very successful robotic platform, called Ladybug, is presented by \citet{Underwood2017Efficient}. The Ladybug robot, among other sensors and navigation equipment, is also equipped with an HSI device (Resonon Pika II, VNIR: 390-887nm) built for low horticultural crops. The robot AGV is a straddle model, where the illumination constancy is granted through the use of a stroboscopic light, in order to do not interfere with the rest of the optical sensors. In addition, calibration pads have been placed both on the field and on the robot, on the periphery of the HSI device's field of view (FOV). The camera has been used to compute only the Normalized Difference Vegetation Index (NDVI), as the whole application was built to gather phenotyping information such as the plant height. The use of straddle systems is one of the most promising strategies for the application of HSI in the fields, however the calibration challenges remain, and the use for taller crops is unpractical. 

In preparation for robotic use, \citet{Kolapesa2023Estimation} perform a very thorough experimentation using VIS-SWIR contact spectrometer (400-2500nm) to estimate SSC in four wine grapes varieties, namely Chardonnay, Malagouzia, Sauvignon-Blanc and Syrah. Contact probe spectrometers have a similar output to HSI imaging devices, but are meant for precise point sampling. These devices provide an internal illumination source and have to be carried around with a handle, or in a backpack. The cost is still higher than RGB cameras, and more complex to integrate into a robotic platform due to the delicate physical contact requirement. The authors analyze the contribution of different wavelengths to the estimation algorithms in depth and apply various algorithms for the prediction. Across the proposed models -- Partial Least Square Regression (PLSR), Random Forest (RF), Convolutional Neural Networks (CNN), and Support Vector Machines (SVM) -- and four grape varieties they obtain an RMSE ranging from 1.70 to 2.20 °Brix. 

In a follow-up study (\citet{Tsakiridis2023In-situ}), in addition to the spectrometer, the authors introduce the in-field estimation of SSC with the use of a snapshot HSI camera (the Cubert FireflEYE V185, VNIR: 400-1000nm). In a different fashion from other works the main contribution of this work is a method to address the problem of illuminant estimation without the use of a reference tile before each acquisition. Instead they propose to learn the illuminant normalization through the use of an auto-encoder. The ground truth for the SSC estimation is collected with an analog refractometer, sampling the °Brix of a single berry for each bunch. The solution of using a deep neural network to estimate the illuminant distribution bears some similarities with our work as it simplifies the acquisition setup,  but loses the absolute metric accuracy of having a white reference in the scene and still requires periodic re-calibration of the network using an more reliable spectrometer. In addition, the use of a snapshot HSI camera considerably reduces the spatial resolution, with respect to push broom setups. The authors explore four varieties of wine grapes, declaring RMSE values that range from 1.66 to 2.20 °Brix.

While the aforementioned approaches demonstrate the notable successes of HSI in agricultural robotics, they also exhibit several key limitations, including high sensor costs, operational complexity, lengthy post-processing times, and limited adaptability to crops where overarching machinery is impractical, such as table grapes. In contrast, the approach proposed in this work achieves comparable performance without relying on HSI, operates in real-time, and is thus better suited for integration into robotic systems.

\subsection{RGB applications to fruit quality assessment}
Given the limitations of HSI illuminant and color calibration, and the cost of hyperspectral devices, many authors proposed approaches based on standard RGB sensors. 
\citet{Bazinas2022Non-Destructive} have explored the use of color imaging combined with advanced numerical techniques, such as interval numbers (INs), and neural network regressors to estimate grape maturity indices (e.g., Total Soluble Solids (TSS), Titratable Acidity (TA), and pH), demonstrating the feasibility of non-destructive and cost-effective methods for integration into autonomous harvesting systems. The authors use a stereo camera (ZED Mini) to acquire RGB images converted to CIELab color space, under natural daylight. The grape variety is Tempranillo that, as in our case, is a black grape cultivar, that allows for an easier correlation of color and SSC. While similar to our approach, the declared performances are considerably higher, \ie 6.6259 Baumé Mean Squared Error (MSE), which approximately corresponds to an RMSE of 4.63 °Brix. 

\citet{Xia2016Non-invasive} perform a very interesting analytical study on 180 samples of Kyoho grape berries using Samsung industrial RGB camera (SCC-B1011). The authors propose the use of some feature indices, similar in principle to the well-known NDVI, but computed with the standard RGB channels, as the inputs to two estimation methods, Partial Least Squares Regression (PLSR) and Multiple Linear Regression (MLR), obtaining a very low cross-validation error of 0.78 °Brix (RMSE) on a dataset with a standard deviation of 1.37°. The study is performed under laboratory conditions, but it is nonetheless significant for our work, as it shows the effectiveness of data-driven estimations with simple RGB cameras. In this work we show that also in uncontrolled environments it is possible to achieve estimation errors low enough to perform automatic harvesting reliably.

\begin{table*}[h!]  
  \caption{Summary of RGB-based approaches for fruit quality assessment (focus on SSC).}
  \label{tab:rgb_approaches}
  %\begin{adjustbox}{center, max width=\textwidth} % Adjusts table to fit, centers it
  \small % Uncomment if further size reduction is needed
  \setlength{\tabcolsep}{4pt} % Reduce inter-column space slightly
  %\resizebox{\textwidth}{!}{
  \begin{tabular}{@{} >{\RaggedRight}p{2.4cm} >{\RaggedRight}p{2.0cm} >{\RaggedRight}p{2.2cm} >{\RaggedRight}p{2cm} >{\RaggedRight}p{4.8cm} >{\RaggedRight}p{3.5cm} @{}}
    \toprule
    \textbf{Reference} & \textbf{Camera Model} & \textbf{Cultivar(s)} & \textbf{Illumination} & \textbf{Key Method / Features} & \textbf{SSC Performance} \\
    \midrule
    \citet{Bazinas2022Non-Destructive} & ZED Mini 3D & Tempranillo (100 samples of 10 vine trees in multiple dates) & Daylight & Color imaging, Interval Numbers (INs), Neural Network regressors & RMSE 4.63 °Brix \\
    \addlinespace
    \citet{Xia2016Non-invasive} & Samsung SCC-B1011 camera & Kyoho (180 berry samples) & Laboratory conditions & Custom RGB feature indices; PLSR and MLR & RMSE 0.78 °Brix (Cross-Validation). \\
    \addlinespace
    \citet{Cavallo2019Non-destructive} & RGB camera & Italia, Victoria (200 bunches) & Artificial illumination & CIELab color histograms (clustered, high variance colors), Random Forest, background subtraction. & 5-level quality scale. \\
    \bottomrule
  \end{tabular}
  %}
  %\end{adjustbox}
\end{table*}

\citet{Cavallo2019Non-destructive} proposed a non-destructive estimation technique based on an RGB camera (JAI CV-M9GE industrial CCD camera) using random forest predictors trained on color histograms. The histogram was computed in CIELab color space by clustering the colors and then choosing the ones with the highest intrinsic variance. While this is the only study that targets white grapes, the estimation problem was framed as a qualitative color scale measurement. The classification task was performed on a 5-level quality scale, and no explicit SSC measurement was included in the labeling process. However, the harvested bunches provided were nominally above 16° Brix level, that for the table grape food industry is a common quality requirement. The experimental setup involved 400 bunches, evenly divided between Italia and Victoria varieties, shot with artificial illumination, and image features were computed after background subtraction and careful white balancing using color calibration references. While this study cannot be quantitatively compared to ours, it is noteworthy as it shows that even with white grape varieties the correlation between color and SSC can be captured by careful modeling. The main limitation of this work that we are trying to overcome is the work in a controlled environment. For this reason, in this initial study we focus our attention to a black grape variety, as in \citet{Bazinas2022Non-Destructive} and \citet{Xia2016Non-invasive}.
Table~\ref{tab:rgb_approaches} summarizes these RGB-based approaches to SSC estimation in grapes.

This growing body of research on RGB-based estimation of internal crop characteristics highlights both the scientific and practical interest in this approach. However, to the best of our knowledge, existing methods either fall short of achieving the performance levels of HSI imaging or fail to transition effectively from controlled laboratory conditions to field environments with uncontrolled lighting.

This study addresses the challenge of perceptual quality estimation for robotic table grape harvesting, a task that could enable robots to assist or even surpass human workers. We model our approach on the standard harvesting procedure where an agronomist first confirms a target SSC in °Brix using a refractometer. Subsequently, workers harvest individual bunches based on visual cues like color. Although SSC and color are not causally linked, they share latent ripening factors -- such as climate, weather, training practices and use of nitrogen, to name a few  (\citet{Jackson1986FActors}) -- which allows human experts to visually approximate ripeness. Our central goal is to replicate and enhance this expert capability using data-driven methods with low-cost RGB sensors.

Our primary contributions are:

\begin{itemize}
    \item A multi-season, multi-sensor dataset of in-field grape images, labeled with both SSC (°Brix) and agronomic color categories. To the best of our knowledge, this is the first publicly available dataset that combines images with °Brix labeling, thus enabling studies on quality estimation under realistic field conditions.
    \item An analysis of the visual-SSC correlation, including a user study that establishes a baseline for human-level performance in estimating ripeness from images.
    \item The design and validation of two distinct algorithms for real-time, calibration-free operation:
    \begin{enumerate}
        \item[a)] A lightweight, histogram-based method for resource-constrained robotic systems.
        \item[b)] A deep, fully convolutional network that achieves high performance even with limited data.
    \end{enumerate}
\end{itemize}

The remainder of this paper is structured as follows. Section~\ref{sec:methods} details our dataset and proposed estimation methods. Section~\ref{sec:experiments} presents the experimental validation and and Section~\ref{sec:discussion} discusses performance, limitations and computational constraints. Finally, Section~\ref{sec:conclusions} concludes the paper and outlines future research directions.
\section{Materials and Methods} \label{sec:methods}
\begin{figure*}[th]
    \centering
    \includegraphics[width=0.8\textwidth]{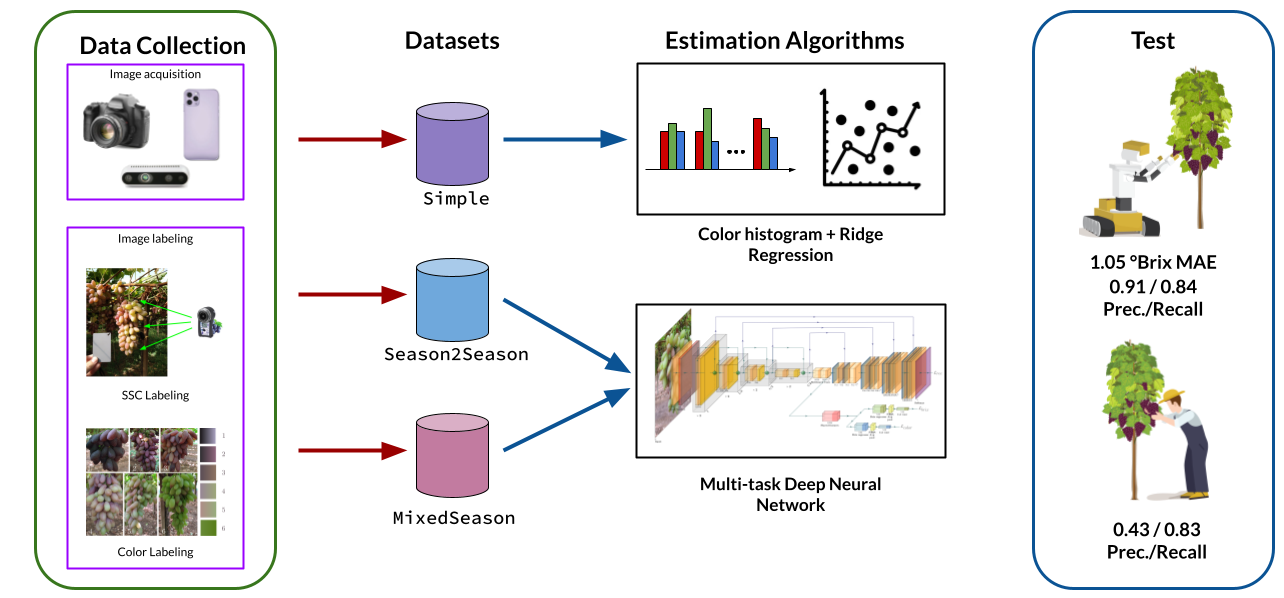}
    \caption{An overview of the experimental workflow presented in this paper. The process moves from left to right: (1) In-field Data Collection using multiple RGB sensors and subsequent labeling for SSC and color quality. (2) Organization of the data into three distinct Datasets to evaluate performance under different conditions. (3) Development and validation of two estimation algorithms: a lightweight histogram-based model and a high-performance multi-task DNN. (4) The Test panel summarizes the key result, outlining our best model's SSC estimation error (MAE) and harvesting decision performance (Precision/Recall) against the human expert baseline. }
    \label{fig:visual_guide}
\end{figure*}

In this Section we detail the data acquisition and labeling processes and the two algorithmic approaches to SSC estimation. Fig, \ref{fig:visual_guide} shows the general overview of the methodology. 

\subsection{Dataset}
In this Section, we describe the data collection and labeling process. 
\subsubsection{Experimental Field} \label{sec:field}
The experimental field is located in southern Lazio, Italy. The vineyard consists of two plots measuring approximately 114 m x 51 m (0.58 ha) and 122 m x 48 m (0.58 ha). 
The vineyards are structured as a traditional trellis system known as \textit{Tendone}, with a wide distance between each plant (3 x 3 m).
All structures are traditionally covered with plastic and netting to protect grapes from rain and hail. 
The plantations are all older than three years, healthy, and in full production, thus representing a typical working condition for the validation of agronomic activities such as fruit harvesting or vine pruning. 
Among the varieties cultivated in this field, Black Pizzutello (\textit{Pizzutello nero}), a distinct variety from the Lazio region, was selected due to its importance for the regional economy and color variability.

\subsubsection{Data collection and labeling} \label{sec:data_labeling}
\begin{figure}[t]
    \centering
    \includegraphics[width=0.8\columnwidth]{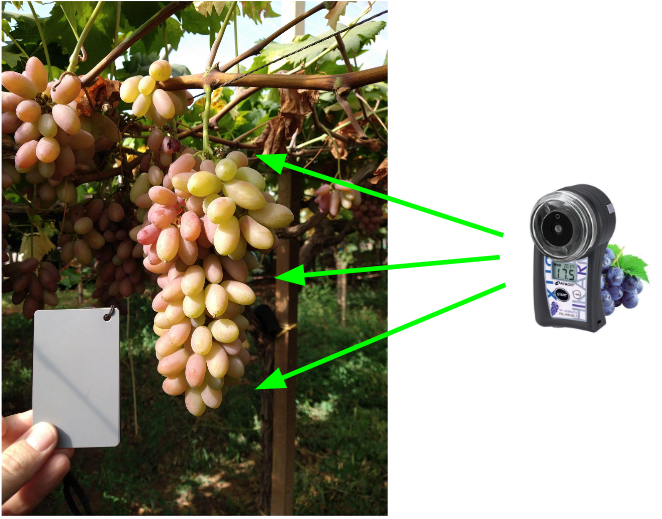}
    \caption{Example of an image collected with a gray card for white balance correction. An agronomist performs on-field measurements by randomly sampling the SSC of berries from the top, middle, and lower parts of the bunch and then averaging the values.}
    \label{fig:brix_sampling}
\end{figure}

\begin{figure} 
    \centering
    \begin{subfigure}[t]{0.195\columnwidth}
        \centering
        \includegraphics[width=\columnwidth]{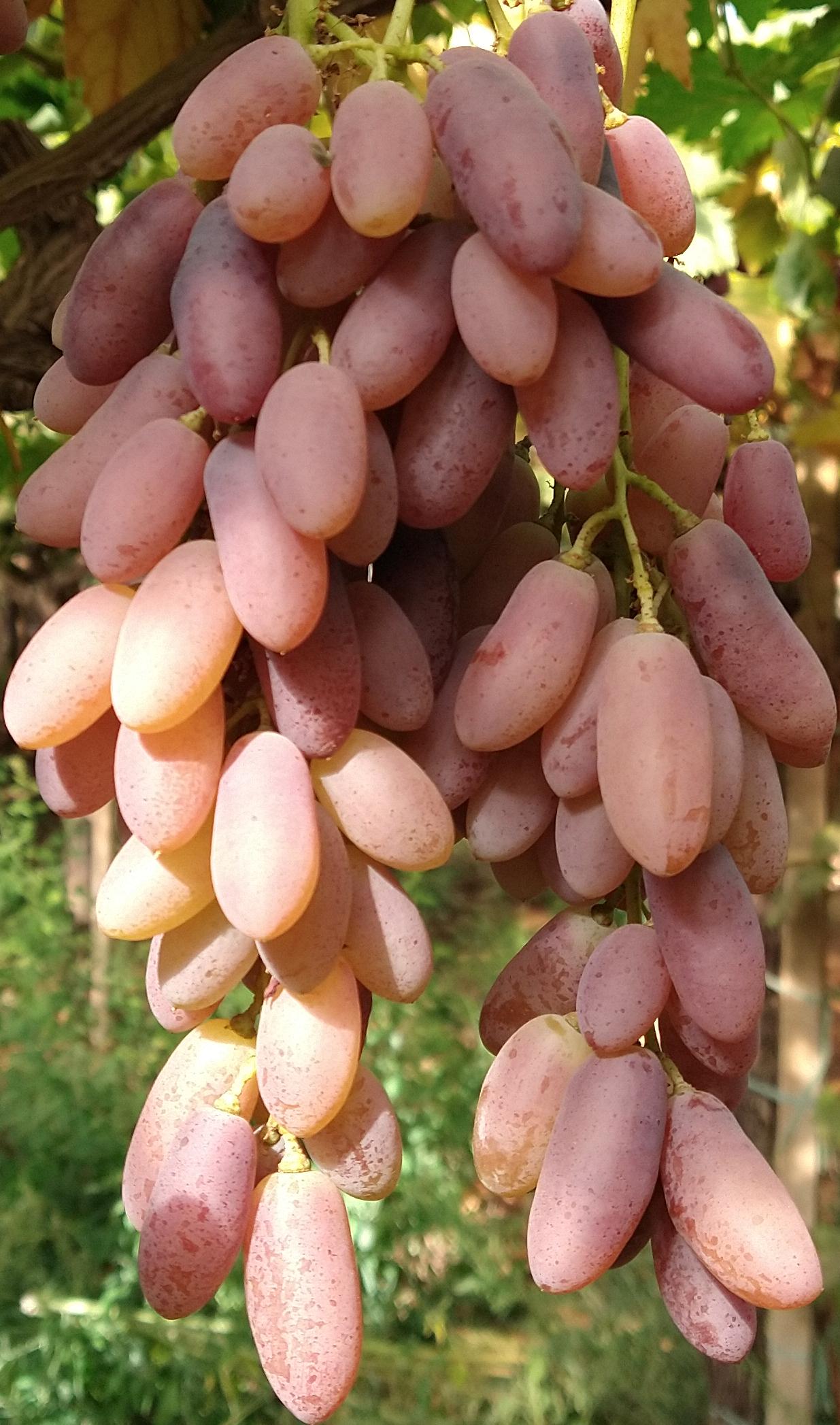}
    \end{subfigure}
    \begin{subfigure}[t]{0.195\columnwidth}
        \centering
        \includegraphics[width=\columnwidth]{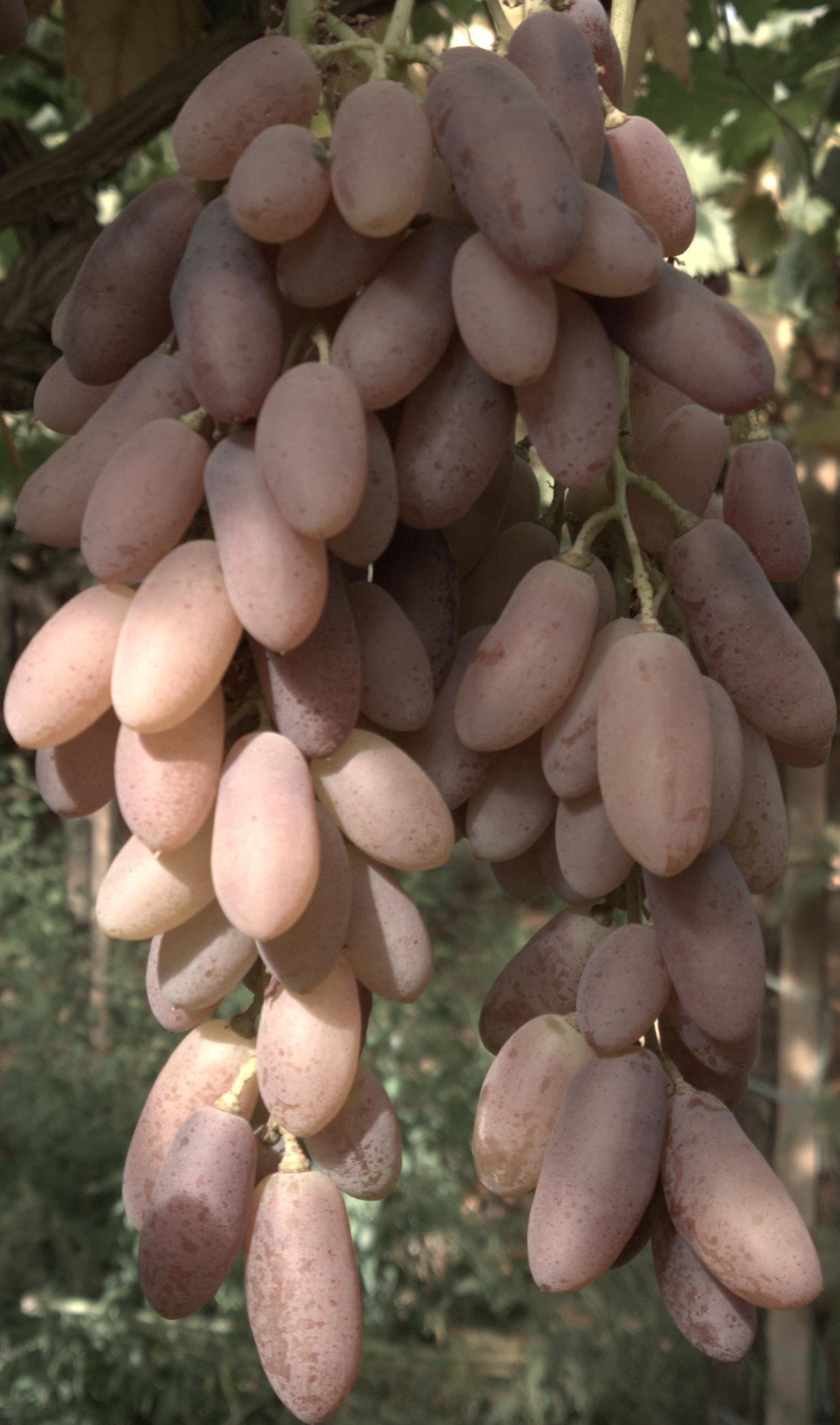}
    \end{subfigure}
    \begin{subfigure}[t]{0.245\columnwidth}
        \centering
        \includegraphics[width=\columnwidth]{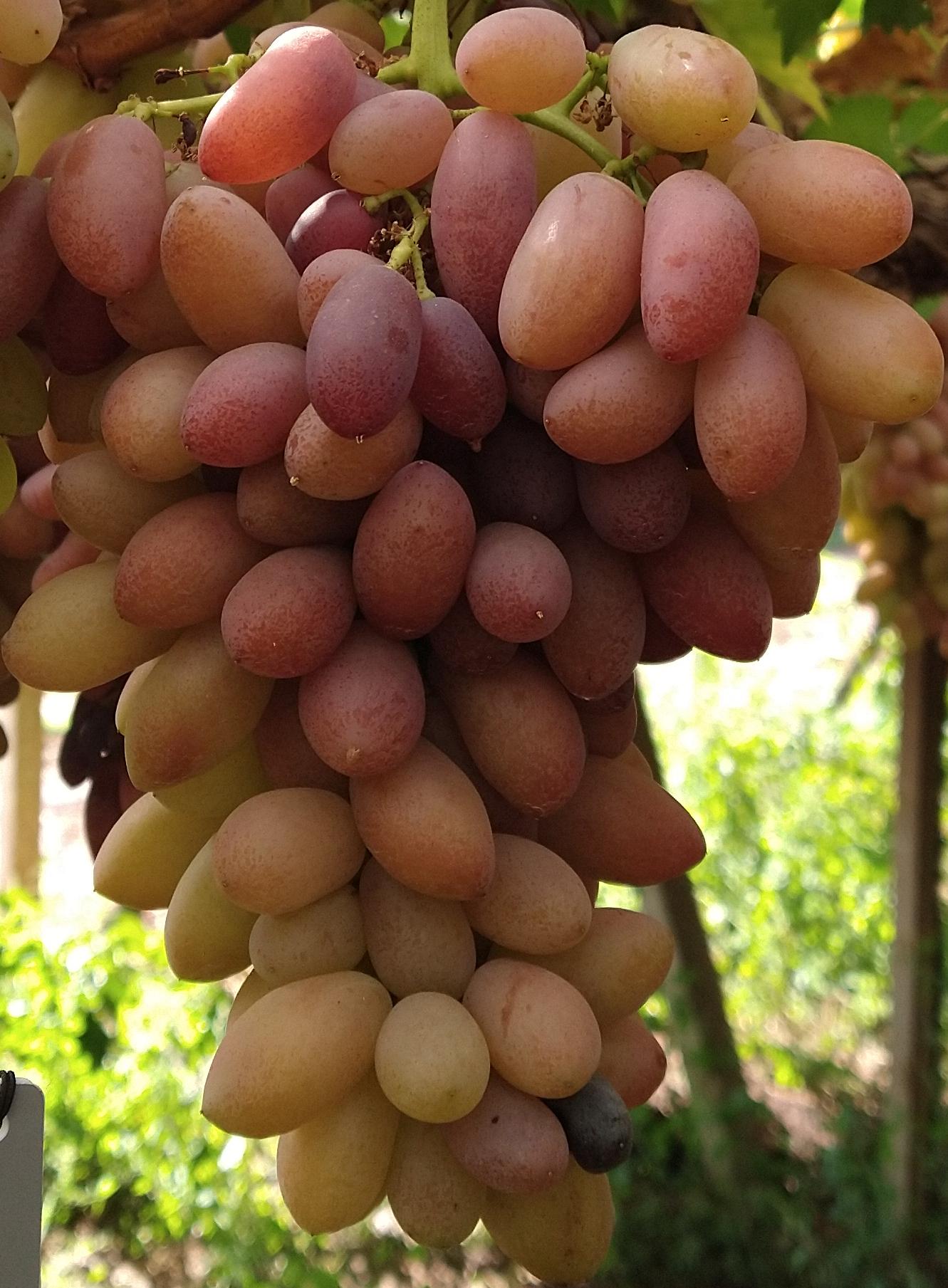}
    \end{subfigure}
    \begin{subfigure}[t]{0.245\columnwidth}
        \centering
        \includegraphics[width=\columnwidth]{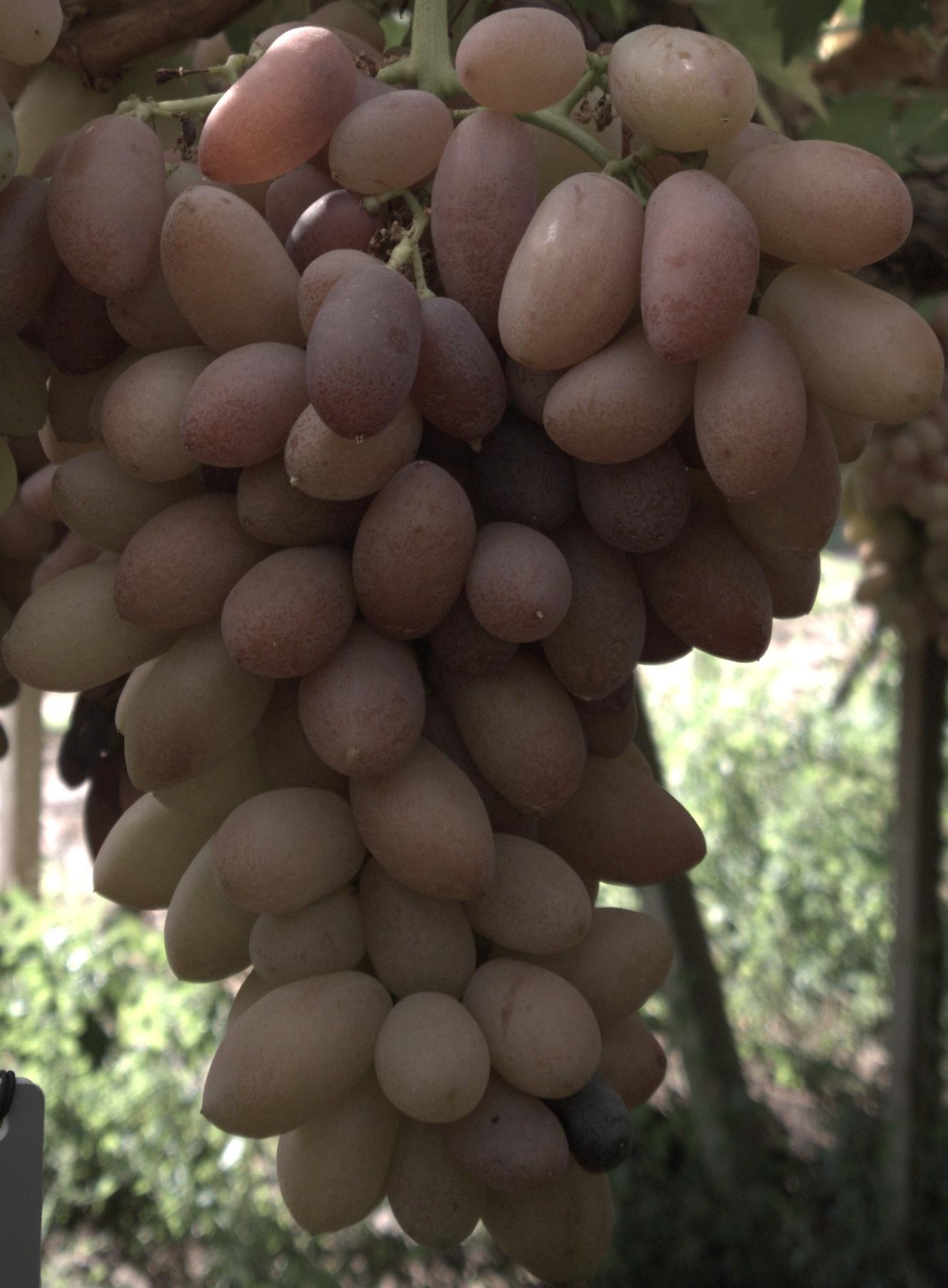}
    \end{subfigure}
    \caption{Some examples of paired JPG and RAW (white-balanced) images collected with the same phone device.}
    \label{fig:std_wb_image_comparison}
\end{figure}

We collected static images during two summer seasons using different camera sensors. Three different devices, across multiple years, have been used to collect the images to represent some of the possible covariate shifts that occur between the collection of the initial data on the field by a farmer or an agronomist and the deployment on a robotic hardware. \\
\textbf{Images capture: }We acquired images using two different smartphones, a Reflex camera, and a Realsense d435i. The smartphone images were captured with a Moto G8 Plus and a Realme GT Master Edition, respectively at $3000\times4000$ and $3448\times4624$ pixel resolution in JPG and raw formats. The Reflex images were captured in raw format at a $3900\times2613$ pixel resolution, while the d435i provided RGB images at a lower $1280\times720$ pixel resolution. When available, raw images were corrected for white balance using a gray card (Fig.~\ref{fig:brix_sampling}), though images without a gray card were still used for comparison. Fig.~\ref{fig:std_wb_image_comparison} shows some examples of images collected with the same smartphone both in JPG and white-balanced RAW format. All images were labeled with bounding boxes around the grape bunches and subsequently cropped. The resulting images vary in size, as some are close-ups while others contain multiple targets. The combination of season, sensor, and gray card presence creates differences summarized in Table~\ref{tab:atomic_splits}. The rows in the table represent atomic splits—groups with the same data distribution—and the variations between them introduce the covariate shifts we aim to study.

\begin{table*}[h!]
    \centering
    \caption{Description of the atomic splits according to the device, image format, gray card presence, and the possibility of having expert ground truth from white-balanced images. }
    \label{tab:atomic_splits}
    \begin{adjustbox}{center}
    \resizebox{\textwidth}{!}{
    \begin{tabular}{||c|c|c|c|c|c|c|c|c|c|c|c|c|c|c||}
    \hline
    \multirow{2}{*}{Dataset Id} & \multirow{2}{*}{Year} & \multicolumn{4}{|c|}{Smartphone} & \multicolumn{2}{|c|}{Reflex} & \multicolumn{3}{|c|}{d435i} & \multirow{2}{*}{Num images} & \multirow{2}{*}{Num samples} & \multirow{2}{*}{Expert GT} & \multirow{2}{*}{Color GT} \\ \cline{3-11}
    & & Type & Jpg & Raw & Card & Raw & Card & RGB & IR & Card &&&& \\
    \hline
    A & 2021 & motog8 & yes & no & no & no & - & no & no & - & 100 & 100 & no & no  \\
    B & 2021 & motog8 & no & yes & yes & yes & yes & no & no & - & 38 & 38 & possible & yes  \\
    C & 2021 & motog8 & yes & yes & yes & yes & yes & no & no & - & 62 & 62 & possible & yes  \\
    D & 2021 & motog8 & yes & yes & yes & yes & yes & no & no & - & 50 & 50 & possible & yes  \\
    E & 2022 & realmi & yes & yes & no & no & - & yes & no & no & 1 & 2 & no & no  \\
    F & 2022 & realmi & yes & yes & yes & no & - & yes & no & yes & 6 & 22 & possible & yes  \\
    G & 2022 & realmi & yes & yes & no & no & - & yes & no & yes & 1 & 2 & no & no  \\
    H & 2022 & realmi & yes & yes & yes & no & - & yes & no & yes & 4 & 19 & possible & yes  \\
    I & 2022 & motog8 & yes & yes & yes & no & - & yes & yes & yes & 8 & 51 & possible & yes  \\
    \hline
    \end{tabular}
    }
    \end{adjustbox}
    
\end{table*}

\begin{figure}
    \centering
    \includegraphics[width=0.9\columnwidth]{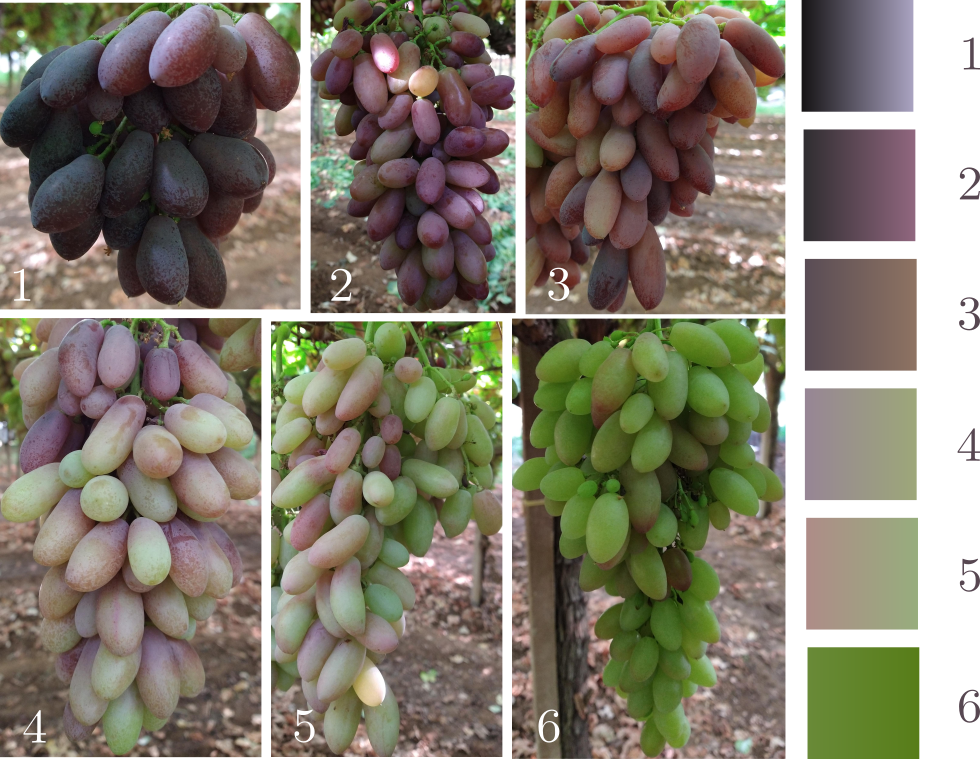}
    \caption{Examples of the six color levels for color classification. A clear distinction between categories may be challenging for an untrained eye. Agronomist classify the color category taking into account both the general color shades of the bunch and the density of highly colored berries. However, our labeling validation tests described in Section~\ref{sec:color_assessment} indicate consistent labeling despite some uncertainty in the classification of neighboring classes.}
    \label{fig:color_chart}
\end{figure}

\textbf{Labeling protocol}: All the images were labeled both for SSC and color. For SSC, we used an Atago Hikari PAL2 digital refractometer, capable of measuring berry sugar content in °Brix by contact without damaging the berry itself. This instrument has a range from 9° to 26°. Typically, depending on the grape variety, an SSC of 16°-18° is considered high enough for harvesting. Since we are trying to assess how well the estimation based on appearance would perform compared to human performance, we adopted a labeling procedure similar to the SSC sampling performed by agronomists. Fig.~\ref{fig:brix_sampling} visually illustrates the process. Every berry in a grape bunch has a different SSC, and the sugar concentration in each berry is also not uniform due to gravity. For this reason, common practice is to sample the SSC three times, from the top, middle, and bottom of the bunch, and then average the results. For consistency, the refractometer was placed on the side of the berry in each measurement.
For color, a categorical classification was required. Based on agronomic production use cases, we defined a color quality chart with the help of experienced agronomists. The color scale goes from 1 to 6, with whole number increments, where 1 represents the best quality and 6 the worst. Usually, values of 1 and 2 are considered high-quality produce, while 3 is still acceptable for some markets. All the other values are considered unripe. This color index accounts for color saturation and distribution across the berries and the bunch. An example of the color variability is given in Fig.~\ref{fig:color_chart}.

\textbf{Experimental Dataset Splits:} Our dataset is built upon a foundation of atomic splits, which are minimal, homogeneous groups of images categorized by season, device, and processing type, as detailed in Table~\ref{tab:atomic_splits}. We combined these atomic splits to create several experimental datasets designed to evaluate model performance in specific scenarios. These combined datasets are distinguished by their image type: standard JPG images (indicated by the \texttt{STD} subscript) or white-balanced images derived from raw captures (indicated by the \texttt{WB} subscript). A summary of these experimental datasets is presented in Table~\ref{tab:mixed_splits}. 

The datasets are grouped by their analytical purpose:

\begin{itemize}
    \item The \Sm splits are small, single-season (2021) datasets used for initial, exploratory assessments of our models. They do not have a predefined train/test division.
    \item  The \Ss splits fix the sensor type to smartphones and are designed to evaluate cross-seasonal performance by training on data from one season and testing on the next.
    \item  The \Ms splits represent the most challenging scenario. They are designed to simulate a real-world deployment by training on smartphone and Reflex images (mimicking data collection by agronomists) and testing on d435i images (mimicking a robotic platform). This directly tests the models' robustness to the covariate shift induced by different acquisition devices.
\end{itemize}

To provide a quantitative baseline for our experiments, Table~\ref{tab:dataset_summary} summarizes key statistical properties of these datasets. The table details the size of each train and test split, along with the natural variability of the SSC labels, measured by the mean, standard deviation, and Mean Absolute Deviation (MAD). These statistics will serve as a reference for evaluating the performance of our proposed estimation models.

\begin{table*}[h!]
    \centering
    \caption{Summary of the experimental dataset splits used for model training and evaluation.}
    \label{tab:mixed_splits}
    \begin{adjustbox}{center}
    \resizebox{\textwidth}{!}{
    \begin{tabular}{||c|c|c|c||}
    \hline
    Name & Train set components & Test set components & Description \\
    \hline
    \Smj & \multicolumn{2}{c|}{C+D} & Some jpg images taken with a phone in 2021. \\
    \Smw & \multicolumn{2}{c|}{C+D} & Some jpg images taken with a phone in 2021 white balanced. \\
    \Ss & A+C+D & F+H+I & All jpg images taken with a phone.  \\
    \Ssw & B+C+D & F+H+I & All raw images taken with a phone that have been white-balanced.\\
    \Ssa & B+C+D & F+H+I & All phone and reflex images white-balanced. \\
    %Phone\_d435i\_Std & \multicolumn{2}{c|}{A+C+D+F+H+I} & All phone images not white-balanced, plus all d435i images \\
    %AllWB\_d435i\_Std & \multicolumn{2}{c|}{B+C+D+F+H+I} & All phone and reflex images white-balanced, plus all d435i images \\
    \Msj & A+C+D+H+I & F & All jpg images, keeping for testing only the ones with the most uncertain brix value (around 16). \\
    \Msw & B+C+D+H+I & F & All white balanced images, keeping for test only the most uncertain brix value (around 16). \\
    \hline
    \end{tabular}
    }
    \end{adjustbox}
    
\end{table*}

\begin{table*}[h!]
    \centering
    \caption{Statistical properties of the experimental dataset splits.}
    \label{tab:dataset_summary}
    \begin{adjustbox}{center}
    \resizebox{\textwidth}{!}{
    \begin{tabular}{||l|c|c|c|c|c|l||}
    \hline
    Name & Split & \# samples & SSC mean & SSC std & SSC MAD  & Description \\
    \hline
    \Smj & -  & 112 & 17.19 & 2.43 & 1.99 & Phone images in JPG format captured in the 2021 season \\
    \hline
    \Smw & -  & 112 & 17.19 & 2.43 & 1.99 & Phone images color corrected from RAW captured in the 2021 season \\
    \hline
    \multirow[c]{2}{*}{\Ssj} & tr & 212 & 16.23 & 2.52 & 2.02 & Phone images captured in JPG format during the 2021 season.  \\
     & te & 92  & 18.91 & 2.16 & 1.71 & Phone images captured in JPG format during the 2022 season.  \\
    \hline
    \multirow[c]{2}{*}{\Ssw} & tr & 150 & 17.54 & 2.46 & 2.04 & Phone images captured in RAW format during the 2021 season. \\
     & te & 92  & 18.91 & 2.16 & 1.71 & Phone images captured in RAW format during the 2022 season.\\
    \hline
    \multirow[c]{2}{*}{\Ssa} & tr & 300 & 17.54 & 2.46 & 2.04 & Phone and reflex images in RAW format, 2021 season. \\
     & te & 92  & 18.91 & 2.16 & 1.71 & Phone and reflex images in RAW format, 2022 season. \\
    \hline
    \multirow[c]{2}{*}{\Msj}  & tr & 352 & 17.67 & 2.75 & 2.33 & All jpg images, across all seasons and devices.\\
     & te & 44  & 15.96 & 1.68 & 1.36 & The test set contains images with SSC content close to the threshold (16° Brix).\\
    \hline
    \multirow[c]{2}{*}{\Msw}  & tr & 440 & 18.27 & 1.98 & 1.96 & All white balanced images, across all seasons and devices.\\
     & te & 44  & 15.96 & 1.36 & 1.36 & The test set contains images with SSC content close to the threshold (16° Brix).\\
    \hline
    \end{tabular}
    }
    \end{adjustbox}
\end{table*}

\subsection{SSC Estimation Models}
In this Section, the two algorithmic approaches to SSC estimation are described. 
\subsubsection{Estimation with Histogram Features}\label{sec:hist_features}
\begin{figure}
    \centering
    \includegraphics[width=\linewidth]{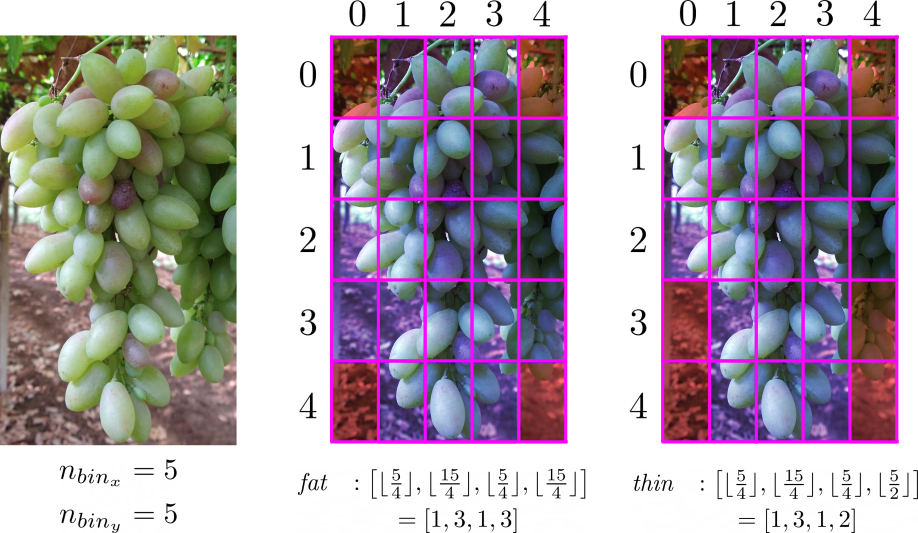}
    \caption{An illustration of the "cross-shaped" feature selection process for histogram-based estimation. An input image is divided into a grid of $n_{bin_x}$ by $n_{bin_y}$ bins ($5\times5$ grid shown). To heuristically remove background pixels, two patterns are tested: a fat pattern that retains a wider central cross-section and a thin pattern that is narrower at the bottom, mimicking the typical shape of a grape bunch. The indices of the kept bins are determined empirically using Equation \ref{eq:cross_limits}. For the $5\times5$ example shown, this results in the index ranges [1,3,1,3] for the fat pattern and [1,3,1,2] for the thin pattern.}
    \label{fig:histogram_features}
\end{figure}

The main use-case for the estimators we design in this work is to be deployed on a robotic platform, embedded edge devices, or smartphone-like devices, thus with limited memory and computational power. For this reason, the first approach is based on classic hand-crafted features, specifically on histograms. RGB channel histograms were chosen as the basic building blocks since the problem largely depends on color intensity and distribution.
To capture distribution differences, we divided each bunch image into a grid of rectangular regions and computed the RGB histogram for each region. A key observation that can be made with this setup is that even with tight bounding boxes from the detector, some regions will often contain background information due to the shape of grape bunches. Adding a background subtraction deep network, such as in \citet{Qin2020U2-Net}, could have been effective to some extent, at the cost of changing the scale of the computational costs. For this reason, we decided to opt for a more straightforward approach. 
To address this, we developed a simple heuristic to remove regions likely to contain background pixels. Based on the empirical observation that grape bunches are typically wider in the middle and narrower at the top and bottom, we retain only a "cross-shaped" selection of regions, as illustrated in Fig.~\ref{fig:histogram_features}. We treat the exact dimensions of this cross-shaped pattern as a hyper-parameter to be optimized. To simplify this exploration, we defined three configurations in Equation \ref{eq:cross_limits}: none (keeping all regions), fat (dropping corner regions), and thin (dropping corners and more bottom regions).

More in detail, the three configurations are automatically defined by the selection of the number of horizontal ($n_{bin_x}$) and vertical ($n_{bin_y}$) divisions (the width and height of the grid in Fig.~\ref{fig:histogram_features}) as follows:
\begin{align}
\mathit{none} &: \left[0, n_{bin_x}-1, 0, n_{bin_y}-1\right] \nonumber \\
\mathit{fat} &: \left[\lfloor\frac{n_{bin_x}}{4}\rfloor, \lfloor\frac{3n_{bin_x}}{4}\rfloor, \lfloor\frac{n_{bin_y}}{4}\rfloor, \lfloor\frac{3n_{bin_y}}{4}\rfloor\right] \label{eq:cross_limits}\\
\mathit{thin} &: \left[\lfloor\frac{n_{bin_x}}{4}\rfloor, \lfloor\frac{3n_{bin_x}}{4}\rfloor, \lfloor\frac{n_{bin_y}}{4}\rfloor, \lfloor\frac{n_{bin_y}}{2}\rfloor\right] \nonumber
\end{align}

Averaging the RGB values on each cell and concatenating them produces a final feature vector with length of $3 \times n_{bin_x} \times n_{bin_y}$. In our experimental validation, We considered also the exploration of the effects of the initial image size and the use of Hue-Saturation-Value (HSV) color space. Given the small number of images, to avoid overfitting, we also considered $L2$ regularization. We use Ridge Regression as the estimation algorithm, since it is the simplest interpretable regression algorithm that allows for regularization. Clearly, other more sophisticated models -- such as Support Vector Machines of Random Forests -- can be used for better estimation results, but in this work we are interested in exploring the characteristics of the features and not only the peak performances.  

\subsubsection{Estimation with Multi-Task Learning DNN Architecture}
The SSC and color distribution can be viewed as conditioned on the environmental and agronomic parameters of the field we are considering. While both the color and the SSC depend on the internal chemical state of the berries, this state itself is influenced by macroscopic environmental and managerial factors such as growing conditions, training practices, and nutrient availability. The overall appearance of the grape bunches is also affected by these factors, which we assume largely homogeneous with respect to key management practices and environmental conditions. Nevertheless, our proposed methods are designed to be robust to the inherent noise and micro-variations arising from individual plant differences, which are expected in any real-world field data. 

\begin{figure*}[h!]
    \centering
    \includegraphics[width=\textwidth]{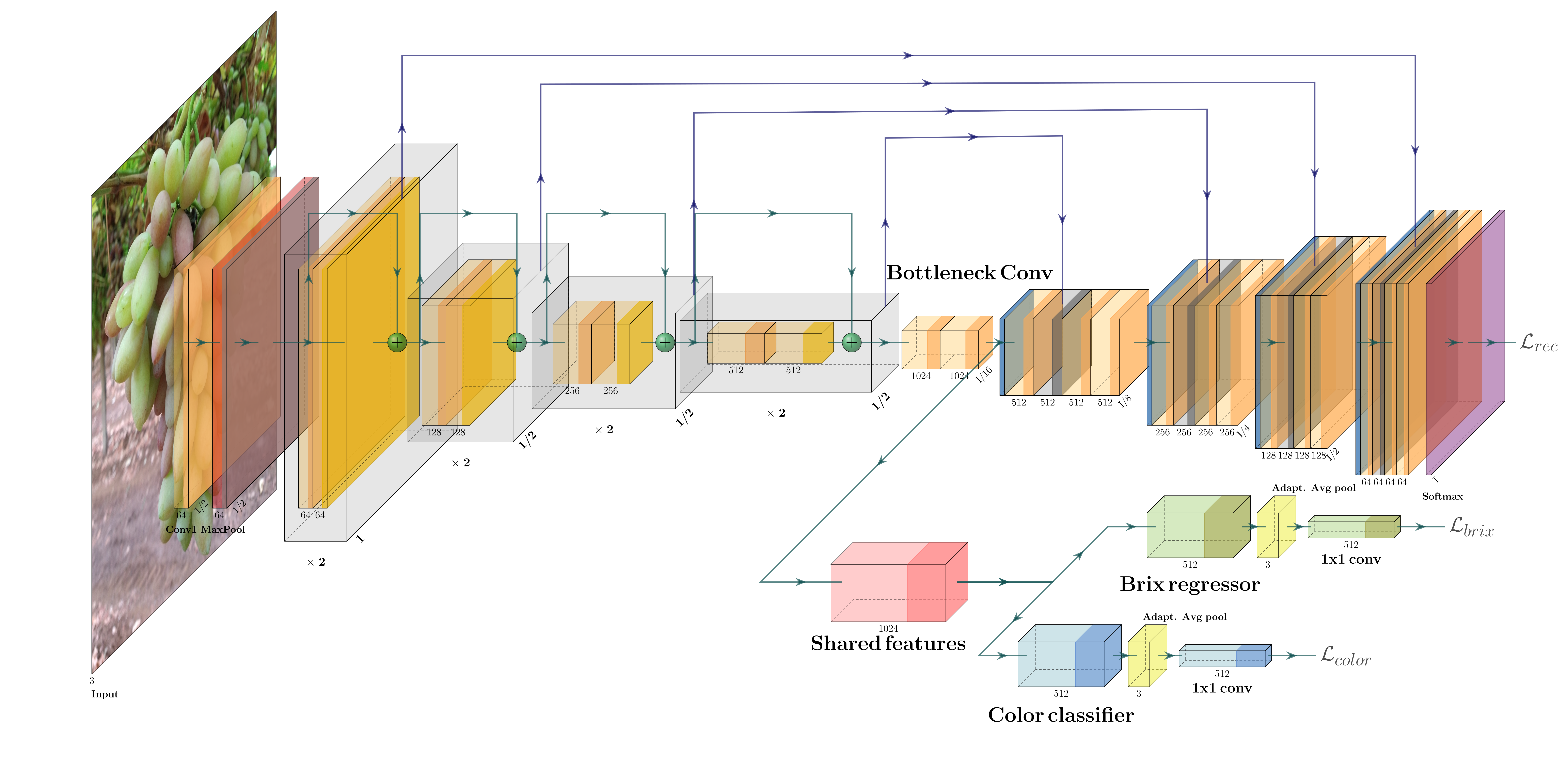}
    \caption{The multi-task network architecture used for SSC and color estimation. The model is based on a U-Net (\citet{Ronneberger2015U-Net}) with a ResNet18 (\citet{he2016deep}) backbone, which acts as an auto-encoder to learn a compact image representation (outputting the reconstruction loss, $\mathcal{L}_{rec}$). Features from the \texttt{Bottleneck Conv} layer are passed to a block of \texttt{Shared features}. From here, the network splits into two parallel heads: a \texttt{Brix regressor} for SSC estimation (outputting $\mathcal{L}_{brix}$) and a \texttt{Color classifier} for quality grading (outputting $\mathcal{L}_{color}$). (Diagram generated with PlotNeuralNet \cite{haris_iqbal_2018_2526396}).}
    \label{fig:dnn}
\end{figure*}

The architecture of the multi-task network, illustrated in Fig.~\ref{fig:dnn}, reflects this reasoning. To distill a feature vector, we employed an Auto-Encoder (AE) architecture, instead of using only a feature extraction backbone alone. The encoder component captures common features in appearance variations that may correlate with the internal chemical state of the grapes. These shared features are then leveraged by two separate prediction heads to estimate the color class and SSC in parallel. The reason is that we are in a data-scarcity scenario, so having the additional reconstruction loss of the auto-encoder, focusing on appearance, helps in strengthening the back-propagation signal and reduces over-fitting. Our auto-encoder is a UNet \citep{Ronneberger2015U-Net} with a ResNet18 \citep{he2016deep} backbone. The two heads share a common "neck," a CNN with 512 output channels, ensuring part of the representation is shared between the tasks. Each head has a symmetrical structure, comprising three CNN blocks that progressively reduce the feature size, followed by adaptive average pooling \citet{zhou2016learning} and a 1x1 convolution \citet{lin2014networknetwork} for the final estimation. In general, UNet-like architectures alone are not suitable for reconstruction problems, since the skip connections would cause the network to simply learn an identity function. Our architecture avoids this problem, by using the bottleneck feature output as the input for the feature extraction section. From those features, a convolutional block is used to extract the features that are then fed to the task-specific heads. The two heads have the same structure, but in the case of color estimation, the outputs are logits for the multi-class problem, while for the SSC estimation are the actual estimations. 

The loss for the architecture is a classical compound loss that combines the reconstruction loss of the AE (pixel-wise MSE), with the addition of a Mean Absolute Error (MAE) loss for the SSC estimate, and a Cross-Entropy loss for the color classes: 
\begin{equation} \label{eq:multi_loss}
    \mathcal{L}_{MT} = \mathcal{L}_{rec} + \alpha \mathcal{L}_{color} + \beta \mathcal{L}_{brix}
\end{equation}
where $\alpha$ and $\beta$ are weighting factors to balance the different contributions of the three losses, whose values have been cross-validated together with other hyper-parameters.

To optimize the hyper-parameters of our multi-task DNN, we developed a custom scoring metric designed to balance statistical accuracy with agronomic needs. We call this the Brix-Color-Hinge (BCH) metric, as it integrates the regression error for °Brix, the classification accuracy for color, and includes hinge-like penalties for predictions that fall below minimum quality thresholds. This approach ensures that the selected model is not only accurate but also practical for harvesting decisions. The detailed mathematical formulation of the BCH metric is provided in \ref{app:hyper-params-validation}.
\section{Experiments} \label{sec:experiments}
This Section describes the experiments of the two proposed algorithmic solutions to SSC estimation. In order to give perspective to these results, some experiments were performed to set a human performance baseline. For this reason, the Section starts with the explanation on how human performances were tested in this scenario (Section~\ref{sec:exploratory}), then proceeds with the actual experiments of the histogram feature model (Section~\ref{sec:hist_experiments}) and multi-task DNN model (\ref{sec:multi_experiments}). Both algorithmic solutions have been initially developed and trained locally on a laptop equipped with a AMD Ryzen 9 5980HX Mobile CPU and a NVIDIA GeForce RTX 3080 GPU with 16GB of VRAM, running Ubuntu 22.04, while the hyper-parameter optimization for all models have been run on a NVIDIA DGX workstation equipped with A100 GPUs. All the code has been written in \texttt{python3}.
The data and code for all the experiments can be found on the accompanying GitHub repository\footnote{\url{https://github.com/Smart-Agri-DIAG/robots-tasting-grapes}}.

\subsection{Human Performance Baselines} \label{sec:exploratory}

\subsubsection{Human Evaluated SSC and Color Correlation} \label{sec:human_evaluation_of_ssc}

\begin{figure}[ht]
    \centering
    \includegraphics[width=\columnwidth]{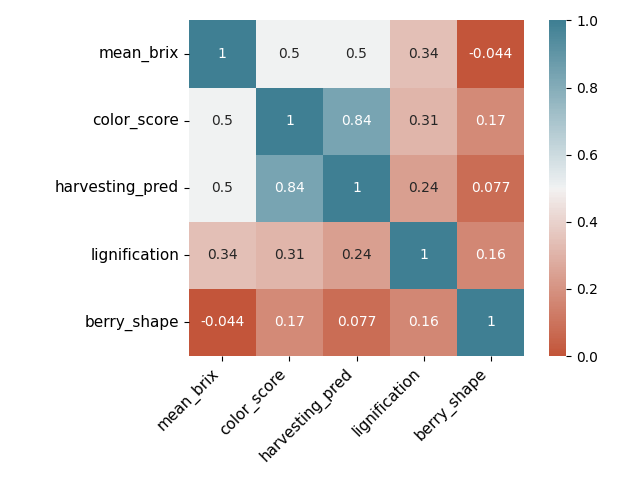}
    \caption{Correlation matrix of key parameters from the human evaluation study on 100 grape images. The matrix shows the relationship between the measured \texttt{mean\_brix} and the scores assigned by experts for \texttt{color\_score}, \texttt{lignification}, and \texttt{berry\_shape}. The strong correlation of 0.505 between \texttt{mean\_brix} and \texttt{color\_score} is the key finding that validates using visual appearance for ripeness estimation.}
    \label{fig:corr_matrix}
\end{figure}

 To establish an initial human performance baseline and assess experts' ability to estimate ripeness from appearance, we performed a specific test. We presented $100$ grape images to a number of agronomists with experience in table grape ripeness estimation. For these images, a °Brix ground truth was collected with the technique described in Fig.~\ref{fig:brix_sampling}. 
 We used the JPG smartphone images collected during the first data collection session (24 September 2021) for Pizzutello Nero. The images were presented without further context, and the experts were asked to assign a numerical value, on a scale from 1-to-10, to the degree of color, peduncle lignification stage, and berry shape, respectively indicated in the following as \texttt{color\_score}, \texttt{lignification}, and \texttt{berry\_shape}. Note that this 1-to-10 color scale differs from the 1-to-6 scale defined in Section~\ref{sec:data_labeling}; This experiment utilized a decimal scale for ease of human interpretation. To create a binary harvesting prediction from this data, we used the agronomists' \texttt{color\_score} as a proxy for their final decision, applying a threshold of 6. We called the thresholded prediction \texttt{harvesting\_pred}. We then measured the correlation between all three visual estimates and the measured \texttt{mean\_brix} to validate our core hypothesis. The correlation scores are shown in Fig.~\ref{fig:corr_matrix}.

The correlation score between the average °Brix (\texttt{mean\_brix}) and color evaluation (\texttt{color\_score}) is $0.505$, with $p<0.001$ and $CI = [0.342, 0.638]$, the correlation between \texttt{mean\_brix} and lignification (\texttt{lignification}) is $0.337$, $p<0.001$ and $CI = [0.151, 0.501]$, while the correlation with berry shape (\texttt{berry\_shape}) score is considerably lower, at $-0.044$, which is below any significance threshold. 

The observed correlation between human expert visual color assessment and actual measured °Brix values suggests that a data-driven algorithm could achieve human-level accuracy in estimating grape quality for harvesting decisions.

This experiments also helps in setting a human performance baseline. Given that a °Brix value greater than $17$ is considered sufficient for harvesting, we use the thresholded \texttt{mean\_brix} as the harvesting ground truth label, comparing it to \texttt{harvesting\_pred}. The precision and recall values were $0.43$ and $0.83$, respectively, translating to a false discovery rate (Type I error rate) of $0.57$ and a false negative rate (Type II error rate) of $0.17$. These values will be compared with similar scores computed with the proposed algorithms in the following Sections.

As a final note to this analysis, it is important to note that in the case of harvesting, Type I errors (false positives) carry a higher cost than Type II errors. This is because ripe bunches left on the vine will eventually be harvested, whereas unripe harvested bunches must be discarded. By collecting table grape bunches with a color class above the threshold of $6$ (the definition for the \texttt{harvesting\_pred} attribute) a significant number of unripe bunches are collected due to a mismatch between °Brix and color. This raises the question of whether increasing the threshold for \texttt{harvesting\_pred} could trade Type I errors for Type II errors. However, computing the precision and recall scores for a \texttt{harvesting\_pred} value obtained with a threshold of $7$ yields $0.35$ and $0.37$, respectively, indicating an increase in both error types. This test shows that, while °Brix and color are correlated across the broad range of the color scale used in this experiment, at the same time, the relationship is not strictly linear. There are a considerable number of both insufficiently colored bunches with sufficient °Brix and highly colored bunches with insufficient °Brix. The high number of bunches in a borderline situation that the experts miss-classify shows that training an appearance-based estimator is a non-trivial task.

\subsubsection{Color Constancy effect on Human Evaluation of Color} \label{sec:color_assessment}
As part of our exploratory analysis, we investigated whether the camera's automatic post-processing (standard JPG) versus a color-corrected image (white-balanced RAW) would significantly affect a human expert's color evaluation. To test this, we conducted a study where agronomists labeled 50 pairs of images (one standard, one white-balanced) in a randomized sequence. A statistical analysis (z-test) of the assigned labels revealed no significant difference between the two groups (p = 0.17). This null finding confirmed that for human experts, the two image types were perceptually similar and justified our subsequent approach of testing our computer vision models on both standard and white-balanced images to evaluate their robustness.

\subsection{Histogram-based Model Evaluation} \label{sec:hist_experiments}
To establish a performance baseline for a computationally inexpensive approach, we evaluated the histogram-based feature model using Ridge Regression. The experiments were conducted on the \texttt{Simple} dataset splits, which contain 112 samples. For this analysis, the data was divided into a training set (89 samples) and a final test set (23 samples). The training images were augmented with horizontal flips, resulting in a total of 178 training images. As previously mentioned, both standard JPG and white-balanced images were evaluated to assess the influence of post-processing on the feature-based algorithms. Model performance and hyper-parameters were then assessed using a 5-fold cross-validation procedure on the training set.
The model's performance was optimized by conducting a grid search over its key hyper-parameters. These included the number of horizontal and vertical histogram bins ($n_{bin_x}$, $n_{bin_y}$), the resulting feature vector length (\texttt{feature\_len}), the choice of color space (RGB, HSV, or CIE-Lab), and the regularization strength ($\lambda$). The full list of parameters and their search ranges is detailed in \ref{app:hist_experiments}.
The best-performing model achieved a validation MAE of $1.46$ °Brix. This result represents a notable $\mathbf{27\%}$ improvement over the statistical baseline for this dataset, which had a Mean Absolute Deviation (MAD) of $1.99$ °Brix.

\begin{figure*}
    \centering
    \begin{subfigure}[t]{0.45\textwidth}
        \centering
        \includegraphics[width=\textwidth]{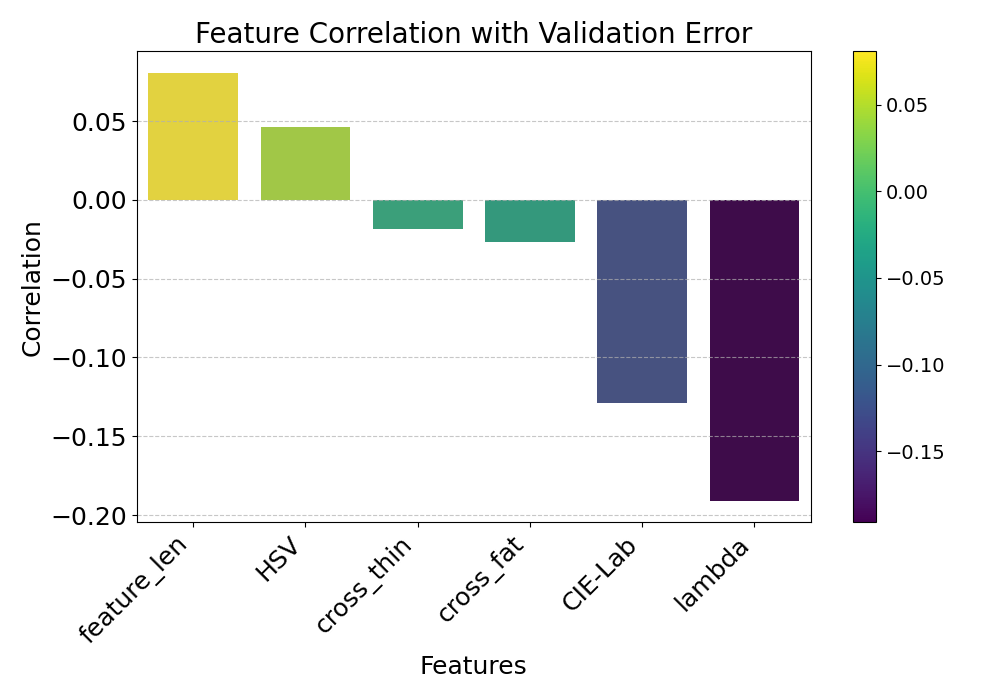}
        \caption{}
        \label{fig:correlation}
    \end{subfigure}
    \hfill % Adds horizontal space between the figures
    \begin{subfigure}[t]{0.45\textwidth}
        \centering
        \includegraphics[width=\textwidth]{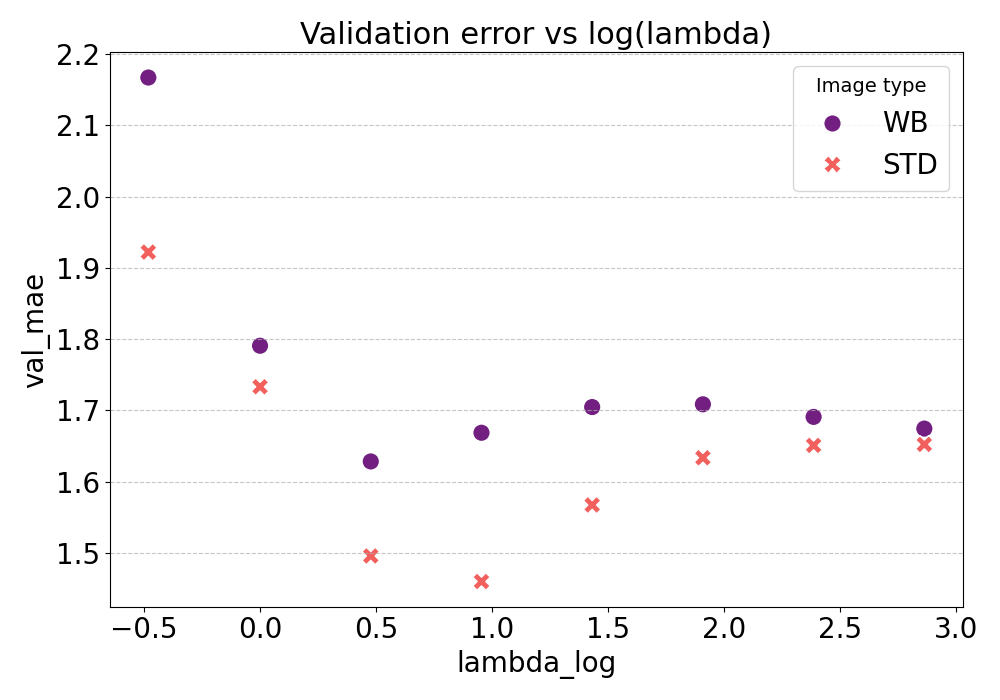}
        \caption{} % Add your caption here
        \label{fig:hist_best_run}
    \end{subfigure}
    \caption{a) Correlation between validation error (MAE) and hyper-parameters for the top 10\% performing runs. This plot highlights the direction (positive or negative) of the correlation. Parameters controlling model complexity, such as \texttt{feature\_len} and \texttt{lambda} (regularization strength), have the strongest correlation with performance. b) Test error for models with hyper-parameters identical to the best model, but with varying \texttt{lambda} values (plotted on a logarithmic scale) and color correction. This plot isolates the effect of regularization and demonstrates that, for this model, standard JPG images consistently yield better performance than their color-corrected counterparts.}
\end{figure*}

The analysis of the hyper-parameter search revealed key insights into the model's behavior. As shown in the feature correlation plot (Fig.~\ref{fig:correlation}), the regularization strength ($\lambda$) and parameters affecting the \texttt{feature\_len} had the most significant impact on performance. The analysis also highlighted the consistent benefit of using an alternative color space, with both HSV and CIE-Lab generally outperforming standard RGB.

Furthermore, we investigated the effect of using standard JPG images versus color-corrected, white-balanced (WB) images. Fig.~\ref{fig:hist_best_run} shows that models trained on the standard JPG images consistently achieved a lower validation error than their counterparts trained on white-balanced images with identical hyper-parameters. This suggests that for this feature-based approach, the camera's default processing pipeline may contain beneficial information, and the extra step of color correction is not necessary.

\subsection{Multi-task DNN Model Evaluation} \label{sec:multi_experiments}

We evaluated the multi-task DNN model on the more challenging dataset splits to assess its performance under realistic covariate shifts. The \Ss splits were used to test cross-seasonal robustness on a consistent device, while the \Ms splits -- with a test set entirely composed by d435i images -- were used to test cross-device generalization, which simulates a real-world deployment scenario. The models were trained according to the training-test splits described in Table~\ref{tab:dataset_summary}. During training, in addition to the usual metrics and the Validation BCH score used for cross-validation, we collected the harvesting Precision, Recall and $F1$ score with the same reasoning used for the human evaluators, \ie by thresholding the SSC prediction and ground truth labels against a threshold of $17$ °Brix,  and comparing them.
The model demonstrated strong performance and generalization capabilities. On the cross-season test set (\Ssj), the best model achieved a MAE of 1.55 °Brix. Performance was even stronger on the more challenging cross-device test set (\Msj), achieving a  MAE of 1.05 °Brix. These results significantly outperform their respective statistical baselines (1.71 and 1.36 °Brix MAD, also in Table~\ref{tab:dataset_summary}), showcasing the model's ability to learn robust visual features for SSC estimation.
These top-performing models were identified through an extensive hyper-parameter search, guided by our custom Brix-Color-Hinge (BCH) metric (the formulation is detailed in \ref{app:hyper-params-validation}).  We observed a high correlation between the minima of the BCH metric and  maxima of F1 score on binary harvesting predictions.  This means that, while during training the two branches could reach better performance individually, usually these were not reflected in a better harvesting estimation. Therefore, the BCH metric effectively allowed to select a subset of top performing models where the double task allowed for better predictions. 

An important observation from the multi-task training process is that the optimal performance for SSC estimation (regression) and color quality (classification) were not always reached at the same training epoch. This highlights an inherent trade-off when using a single feature extractor for multiple correlated but distinct tasks. This also implies that, for a specific application where the top performance of either SSC or color estimation is required, one could select different model checkpoints depending on whether the primary goal is maximizing SSC accuracy or color classification accuracy.

\begin{figure}[t]
    \centering
        \includegraphics[width=\columnwidth]{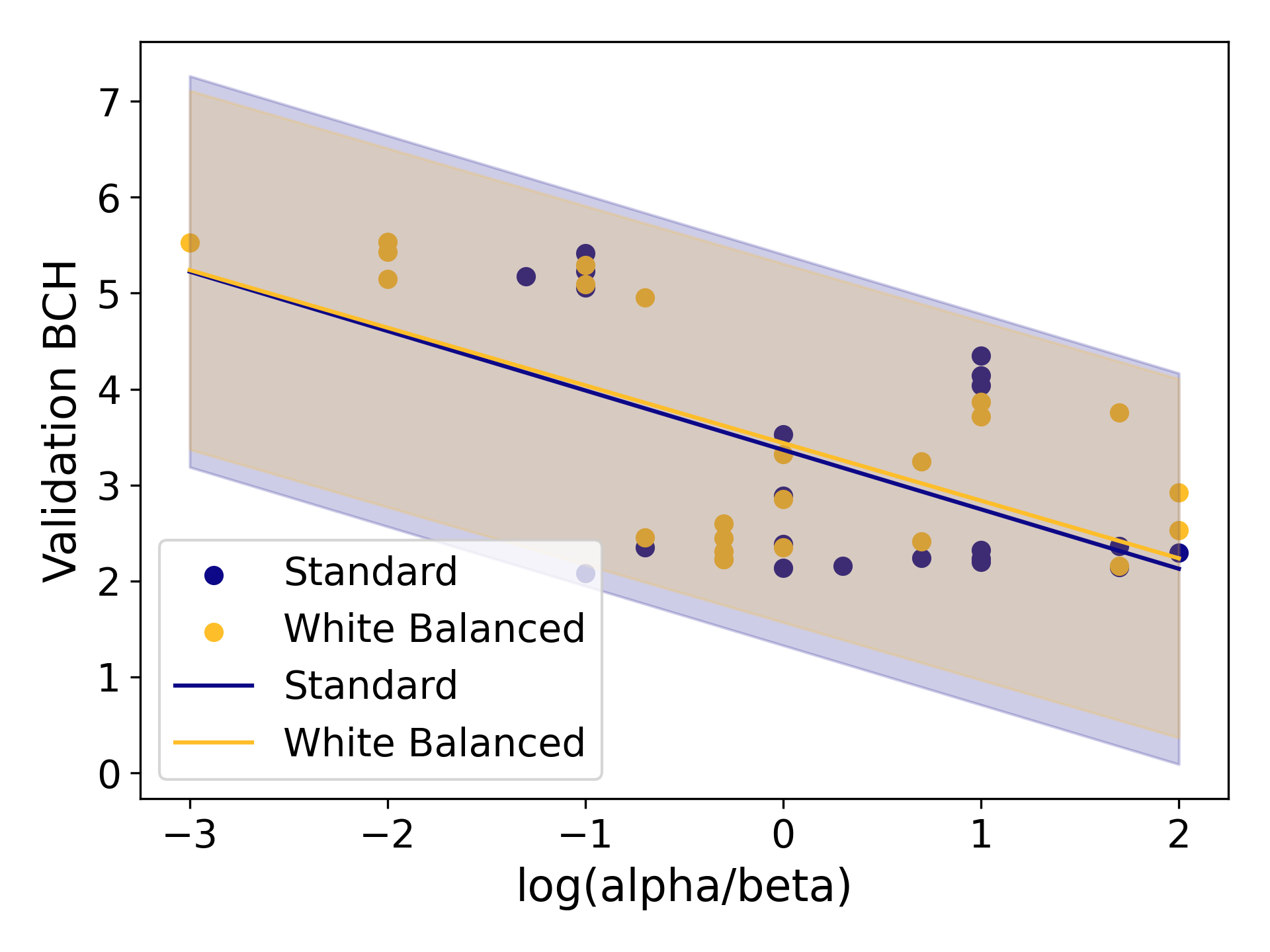}
    \caption{Linear regression analysis of runs with hyper-parameters similar to the top 20 runs on the \Msj and \Msw datasets. The shaded area represents the $95\%$ confidence interval. While the models trained on color-corrected samples present slightly higher values on average, the smaller difference compared to Fig.~\ref{fig:hist_best_run} suggests no practical advantage in using standard images over the white-balanced ones in this context.}
    \label{fig:best_MS_vs_MS_WB_regression}
\end{figure}

Finally, we analyzed the impact of color correction on the DNN model's performance, similarly to what was done with the histogram features. To isolate this effect, we selected the best-performing models on the standard \Msj dataset and compared them against runs using identical hyper-parameters but trained on the white-balanced \Msw data. Fig.~\ref{fig:best_MS_vs_MS_WB_regression} plots the results of this comparison against the log-ratio of the regularization weights ($\alpha / \beta$). The resulting regression lines show that the performance difference between the two sets of images was minimal, with largely overlapping confidence intervals. This suggests that the deep learning model is robust to these color variations and can learn the relevant features effectively without requiring complex color calibration steps prior to deployment.

\section{Discussion} \label{sec:discussion}

The results of our experiments demonstrate that low-cost, in-field RGB imaging can be a viable alternative to more complex and expensive systems for SSC estimation. In this section, we synthesize the performance of our models, compare them against the established baselines and the state-of-the-art, and discuss the practical implications and limitations of our work.

\subsection{Summary of Performance and Comparison}
\label{sec:discussion_summary}

To provide a holistic view of our findings, Table~\ref{tab:master_comparison} consolidates the performance of our proposed models against the statistical and human baselines, as well as against other non-contact methods in the literature.

\begin{table*}[h!]
    \centering
    \caption{Master comparison of model performance against baselines and state-of-the-art methods.}
    \label{tab:master_comparison}
     \begin{adjustbox}{center}
     \resizebox{\textwidth}{!}{
        \begin{tabular}{l l l c c}
        \hline
        \textbf{Model} & \textbf{Method} & \textbf{Dataset / Conditions\footnotemark[2]} & \textbf{Performance Metric}\footnotemark[3]& \textbf{Result} \\
        \hline
        Statistical Baseline & - & \Ss Test & MAD & 1.71 °Brix  \\
        Statistical Baseline & - & \Ms Test & MAD & 1.36 °Brix  \\
        Human Experts & Human Study & 100 JPG images from the 2021 harvest & Precision / Recall & $0.43 / 0.83$  \\
        \hline
        \textbf{Our Histogram Model} & Color Histograms & \Smj & MAE & 1.46 °Brix  \\
        \textbf{Our DNN Model} & Multi-task DNN & \Ssj & MAE & 1.55 °Brix  \\
        \textbf{Our DNN Model} & Multi-task DNN & \Msw & MAE & \textbf{1.05} °Brix  \\
        \textbf{Our DNN Model} & Multi-task DNN & Harvesting Task Comparison on \Ss & Precision / Recall & $\mathbf{0.91} / \mathbf{0.84}$ \\
        \textbf{Our DNN Model} & Multi-task DNN & Harvesting Task Comparison on \Ms & Precision / Recall & $\mathbf{0.62} / \mathbf{0.83}$ \\
        \hline
        \citet{gutierrez2019on-the-go} & HSI & Tempranillo / In-Field & RMSE & ~1.27 °Brix  \\
        \citet{Bazinas2022Non-Destructive} & RGB + NN & Tempranillo / In-Field & RMSE & ~4.63 °Brix  \\
        \citet{Xia2016Non-invasive} & RGB + MLR  & Kyoho / Laboratory & RMSE & ~0.78 °Brix  \\
        \hline
        \end{tabular}
        }
     \end{adjustbox}
    \end{table*}
    
Our deep learning model clearly learned a meaningful pattern from the images, achieving a best MAE of \textbf{1.55} °Brix on the \Ss dataset and \textbf{1.05} °Brix on the challenging cross-device \Ms dataset. This significantly outperforms the naive statistical baselines of 1.71 and 1.36 °Brix MAD respectively for that same test set. 

To better address how these errors translate to harvesting process performance and to benchmark against practical expertise, we also evaluated our DNN model on the binary harvesting task. As shown in Table~\ref{tab:master_comparison}, our model achieved a precision/recall of $\mathbf{0.91}/\mathbf{0.84}$ on \Ss split and of $\mathbf{0.62}/\mathbf{0.83}$ on the \Ms split. This is compared to the experts' performance of $0.43 / 0.83$. Being able to keep a high recall, while showing a higher precision is particularly valuable in a real-world scenario, as it would reduce the costly error of harvesting unripe grapes. In addition, our experiments consistently showed that the F1 score computed from these Precision and Recall values is highly correlated to the BCH metric (in most runs the maxima of the F1 score correspond to the minima of the BCH metric) suggesting that the model's high performance derives from the effective combination of features related to both color and SSC estimation.

When benchmarked against other visual estimation techniques, our low-cost RGB approach demonstrates highly competitive performance. While direct comparison is challenging due to different grape varieties and conditions, our best MAE of 1.05 °Brix is favorable when compared to reported RMSE values from in-field HSI systems, which range from 1.27 to 2.20 °Brix. It also represents a significant improvement over other in-field RGB methods that report errors closer to 4.63 °Brix RMSE, and approaches the accuracy of methods performed under controlled laboratory conditions.
    
\subsection{Practical Implications and Trade-offs}
\label{sec:discussion_implications}

Our work presents two distinct solutions tailored for different operational needs. The histogram-based model, while less accurate, offers significant advantages in computational efficiency and simplicity. Once trained, the estimation reduces to a dot product between the feature vector and the estimator weights. The most expensive operation in that case is the pixel-wise integral sum needed for the histogram computation, which has a computational complexity of O(N), where N is the number of pixels. However, as it was shown in the experiments, given the low frequency content of the grape image it is possible to work with very low resolution images and drastically reduce N. To summarize, the histogram features' low memory and processing footprint make it suitable for real-time deployment on resource-constrained edge devices or as a first-pass filter on a robotic platform. In contrast, the DNN model provides superior accuracy and generalization, particularly across different devices and seasons, at a greater computational cost. This makes it ideal for applications where performance is paramount, such as a centralized quality control system or deployment on robots with dedicated GPU hardware.

\subsection{Limitations and Future Work}
\label{sec:discussion_limitations}

We acknowledge several limitations that provide avenues for future research. This study was conducted on a single black grape variety, \textit{Pizzutello Nero}. While this is common in literature (see Table~\ref{tab:master_comparison}), to extend it to a white variety some adjustments in the setup are required. In the case of black or red grape varieties, the color gradient can be perceived as a contrast change, while in with the white varieties it is a hue and saturation change. To detect variations of the latter kind, a color calibrated RGB imaging system is needed, or, in alternative, to have a calibration pattern in the scene. This is not impossible to achieve and it is an interesting future development for this work. 
Additionally, while our methods showed robustness to the uncontrolled lighting of the field, further experiments under more extreme weather conditions (e.g., direct bright sunlight vs. heavy clouds) could better quantify the operational domain of the system. Future work will focus on expanding the dataset to include these variations and on exploring multi-sensor fusion techniques to further improve robustness.

\footnotetext[2]{SotA results are from different grape varieties and conditions.}
\footnotetext[3]{{MAE and RMSE are not directly equivalent but are presented for contextual comparison.} }

\section{Conclusions} \label{sec:conclusions}

This work successfully demonstrated the feasibility of estimating Soluble Solid Content (SSC) in grapes using low-cost RGB sensors in challenging, uncontrolled field conditions. By creating a novel multi-season, multi-device dataset of \textit{Pizzutello Nero}  labeled for both SSC and color, we established a framework for developing and validating data-driven estimation models, addressing the practical limitations of expensive laboratory or hyperspectral equipment. While numerous studies explore laboratory equipment and hyperspectral data for SSC estimation through visual sensors, this study is among the few to focus exclusively on the practical scenario of RGB data collected directly in field conditions. The complexity of this task arises from covariate shifts inherent to the problem, including seasonal variations and differences in imaging devices.

Two algorithmic approaches were proposed: a lightweight method using classical handcrafted features, suitable for devices with limited computational power, and a multi-task deep neural network architecture that, while more computationally demanding, effectively leverages correlations between visual appearance (color and shades) and SSC.

Both methods were experimentally validated using field-collected data. Results indicate that the proposed techniques achieve reliable SSC estimation with a reasonable margin compared to the average human estimator. Our multi-task deep neural network achieved a best-performing Mean Absolute Error (MAE) as low as 1.05 °Brix on a challenging cross-device test set. This level of accuracy is highly competitive with reported results from in-field hyperspectral systems and is comparable to our human expert baseline, whose harvesting decisions the model could match with high precision and recall. The demonstrated robustness to seasonal and hardware-induced covariate shifts position our solutions as viable prototypes for low-cost, on-field applications.

To tackle this challenge, a dataset spanning multiple years was collected, incorporating data from various camera devices and annotated for SSC and color quality based on agronomic practices. Human performance in assessing the correlation between color and SSC was evaluated through two user studies, which informed the design of data-driven algorithms for estimating both SSC and color.

To foster further research we release our complete dataset, code, trained models, and training scripts used in this study. Future work will focus on  extending the approaches to green cultivars, and on further studying how low-cost estimators can directly approximate hyperspectral devices in agronomic contexts..

\section*{CRediT authorship contribution statement}
{\bfseries Thomas A. Ciarfuglia} Conceptualization, Methodology, Software, Validation, Formal analysis, Data Curation, Investigation, Resources, Writing - Original Draft, Writing - Review \& Editing, Visualization, Supervision, Project administration, Funding acquisition.
{\bfseries Ionut M. Motoi:} Software, Validation, Data Curation, Writing - Original Draft, Writing - Review \& Editing, Visualization 
{\bfseries Leonardo Saraceni:} Software, Validation, Data Curation
{\bfseries Daniele Nardi:} Funding acquisition, Writing - Review \& Editing 

\section*{Declaration of competing interest}
{\small The authors declare that they have no known competing financial interests or personal relationships that could have appeared to influence the work reported in this paper.} 

\section*{Acknowledgments}
{\small
\euflag \quad This work is part of a project that has received funding from the European Union’s Horizon 2020 research and innovation programme under grant agreement No. 101016906 – Project CANOPIES

\euflag \quad This work has been partially supported by project AGRITECH Spoke 9
- Codice progetto MUR: AGRITECH ”National Research Centre for Agricultural
Technologies” - CUP CN00000022, of the National Recovery and Resilience Plan
(PNRR) financed by the European Union ”Next Generation EU”.

This work has been partially supported by Sapienza University of Rome as part of the work for project \textit{H\&M: Hyperspectral and Multispectral Fruit Sugar Content Estimation for Robot Harvesting Operations in Difficult Environments}, Del. SA n.36/2022.}

\bibliographystyle{apalike}
\bibliography{bibliography}

\clearpage
\appendix

\section{Hyper-parameters Exploration for the \\Histogram-based Estimators} \label{app:hist_experiments}

Following from Section~\ref{sec:hist_experiments}, to allow full reproducibility of the experiments, here we give the details of the hyper-parameter search for the histogram features. 

Table~\ref{tab:hyp_sweep} summarizes the hyper-parameter ranges investigated in this experiment.

\definecolor{LightCyan}{rgb}{0.88,1,1}
\begin{table*}[th]
    \caption{Hyper-parameters ranges for the histogram feature-based models.}
     \label{tab:hyp_sweep}
     \centering
     \resizebox{\textwidth}{!}{
     \begin{tabular}{c|c|p{0.6\linewidth}}
         \textbf{Hyper-parameter} & \textbf{Range} & \textbf{description} \\ 
         \hline
         $n_{bin_x}$ & 4, 8, 16 & number of image columns bins. \\
         \rowcolor{LightCyan}
         $n_{bin_y}$ & 8, 16, 32 & number of image rows bins. \\
         $cross$ & "none", "thin" or "fat" & How many bins to drop for background elimination. See Eq.\ref{eq:cross_limits} for details.\\
         \rowcolor{LightCyan}
         $color\_space$ & "none", "hsv" or "Lab" & Use of CIE-La*b*, HSV or RGB color coordinates. \\
         \multirow{2}{*}{$dataset$} & low\_res, med\_res, high\_res, & \multirow{2}{*}{\shortstack[l]{Image resolution, corresponding to $320\times240$, $640\times480$\\ and $1280\times720$ respectively, in color corrected and standard form.}} \\
         & wb\_low\_res, wb\_med\_res, wb\_high\_res & \\
         \rowcolor{LightCyan}
         $\lambda$ & 0.33, 1, 3, 9, 27, 81, 243, 729 & Ridge Regression regularization parameter. \\
         \hline
     \end{tabular}
     }
     
 \end{table*}
 
\begin{figure}[h]
    \centering
    \includegraphics[width=\columnwidth]{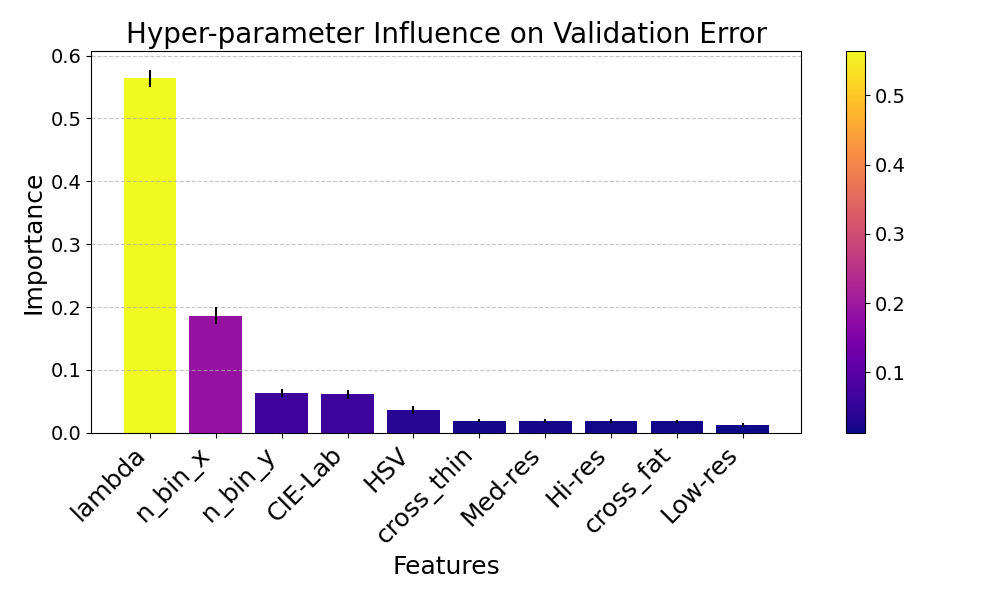}
    \caption{Training a Random Forest using the hyper-parameter values to predict the validation error and then extracting the importance scores is a common way to extract the relative impact of each feature on the final estimator performance. Here we can see that the choice of initial image resolution is not relevant, while the regularization, the number of bins (that influence feature length and flexibility), and the color space have to be carefully selected.}
    \label{fig:hist_importance}
\end{figure}

Given the manageable number of combinations and the low computational cost of each experiments, we performed a full grid search to find the optimal hyper-parameters. To identify the most relevant ones we used the \textit{importance factor}, as described in \citet{Gromping2009Variable} and implemented in the \textit{Weights\&Biases} software suite for experiments tracking \citep{wandb}. Fig.~\ref{fig:hist_importance} gives an example of this analysis, while Table~\ref{tab:hist_best_runs} shows some of the best performing runs. The full results and hyper-parameters for all 3888 experimental runs are publicly available for inspection to ensure full transparency in the \textit{GitHub}\footnotemark[1] and \textit{Weights\&Biases}\footnotemark[5]\footnotetext[5]{\url{https://wandb.ai/sparviero19/CQEM-193_Brix_sweep/reports/CQEM-193-Sweep-results-on-sweep-zgyffmc5--Vmlldzo2Njc3MjA4}} companion sites.

\begin{table}[h]
    \centering
    % \captionsetup{width=.6\textwidth}
    \caption{Top 10 histogram-based models, ranked by validation mean absolute error (Val MAE), along with their corresponding hyper-parameters values.}
    \label{tab:hist_best_runs}
    \resizebox{\columnwidth}{!}{
    \begin{tabular}{rrrrrrrrrr}
    \toprule
    Test MAE & Val MAE & Train MAE & $\lambda$ & $n_{bin_x}$ & $n_{bin_y}$ & $cross$ & $color\_space$ \\
    \midrule
    1.46 & 2.10 & 1.58 & 9 & 16 & 8 & none & HSV \\
    1.51 & 2.13 & 1.55 & 9 & 8 & 16 & none & HSV \\
    1.52 & 2.14 & 1.61 & 3 & 8 & 8 & none & HSV \\
    1.51 & 2.14 & 1.18 & 9 & 8 & 32 & none & HSV \\
    1.52 & 2.16 & 1.62 & 1 & 16 & 8 & thin & CIE-Lab \\
    1.49 & 2.16 & 1.25 & 3 & 16 & 8 & none & HSV & \\
    1.52 & 2.18 & 1.43 & 3 & 16 & 8 & fat & HSV \\
    1.52 & 2.19 & 1.72 & 0.33 & 8 & 8 & thin & CIE-Lab \\
    1.43 & 2.20 & 1.19 & 3 & 8 & 16 & none & HSV \\
    1.49 & 2.33 & 1.22 & 1 & 16 & 8 & thin & HSV \\
    \bottomrule
    \end{tabular}
    }
\end{table}

% --------------- APPENDIX ON CROSS VALIDATION OF MULTI-TASK NETWORK ----------------------- %
\definecolor{LightMagenta}{rgb}{1,0.9,1}
\begin{table*}[t]
\small
     \centering
     \caption{Hyper-parameter descriptions for the Multi-task experiments.}
     \label{tab:multi_hyp_sweep}
     %\resizebox{\textwidth}{!}{
     \begin{tabularx}{\textwidth}{c|X|p{0.5\linewidth}}
         \textbf{Hyper-parameter name} & \textbf{Range} & \textbf{description} \\ 
         \hline
         $\alpha$ and $\beta$ & 1, 10, 50, 100, 500, 1000 & The loss balancing factors (Eq. \ref{eq:multi_loss}). \\
         \rowcolor{LightMagenta}
         $batch\_size$& 32, 64, 128 & Training batch size. \\
         $learning\_rete$ & 0.00001, 0.00005, 0.000001 & Initial learning rate, to be modified by the scheduling strategy. \\
         \rowcolor{LightMagenta}
         $momentum$ & 0.8, 0.9, 0.95, 0.99 & Momentum for the SGD optimizer. \\
         $im\_height$ and $im\_width$ & 64, 128, 256 & The image starting sizes before feature extraction by the auto-encoder. \\
         \rowcolor{LightMagenta}
         $blur\_\sigma_{min}$ and $blur\_\sigma_{max}$ & $\left[0.01, 0.1, 0.5\right]$ and $\left[2.0, 4.0\right]$ & Together they define the range for the Gaussian blur augmentation that is always applied.\\
         $color\_space$ & "none", "hsv" or "Lab" & Use of CIE-La*b*, HSV or RGB color coordinates. \\
         \rowcolor{LightMagenta}
         $one\_cycle$ & $True$ or $False$ & Whether to use of not the one cycle scheduling strategy (\citet{smith2018superconvergencefasttrainingneural}). \\
         $pretrained$ & $True$ or $False$ & Whether the \texttt{resnet18} backbone is pretrained on ImageNet (\citet{he2016deep}) or not. \\
         \rowcolor{LightMagenta}
%         \shortstack[l]{ $dataset\_split$ \\ $\qquad$ \\ $\qquad$ \\}  & \shortstack[l]{{\small \Ssj, \\ \small \Ssw, \\ \small \Ssa, \\ \small \Msj, \Msw}} & \shortstack[l]{The different dataset splits used to test cross-season\\ and cross-device performance}. \\
		 $dataset\_split$  & \small \Ssj \Ssw \Ssa \Msj \Msw & The different dataset splits used to test cross-season and cross-device performance. \\
         \hline
     \end{tabularx}
     %}
 \end{table*}
 
\section{Hyper-parameter Exploration for the Multi-task DNN} \label{app:multi_experiments}
This appendix follows the same principles described in the previous one. Table~\ref{tab:multi_hyp_sweep} gives a summary of the hyper-parameter ranges explored in the sweep. A wide range of hyper-parameters was explored using random search. In this case, the dataset splits used are not to be considered proper hyper-parameters, but as a defining subset of the parameter sweep, since the distribution shifts between them are considerable. They are included in this table for ease of reference. 

\begin{figure}[h]
    \centering
       \includegraphics[width=\linewidth]{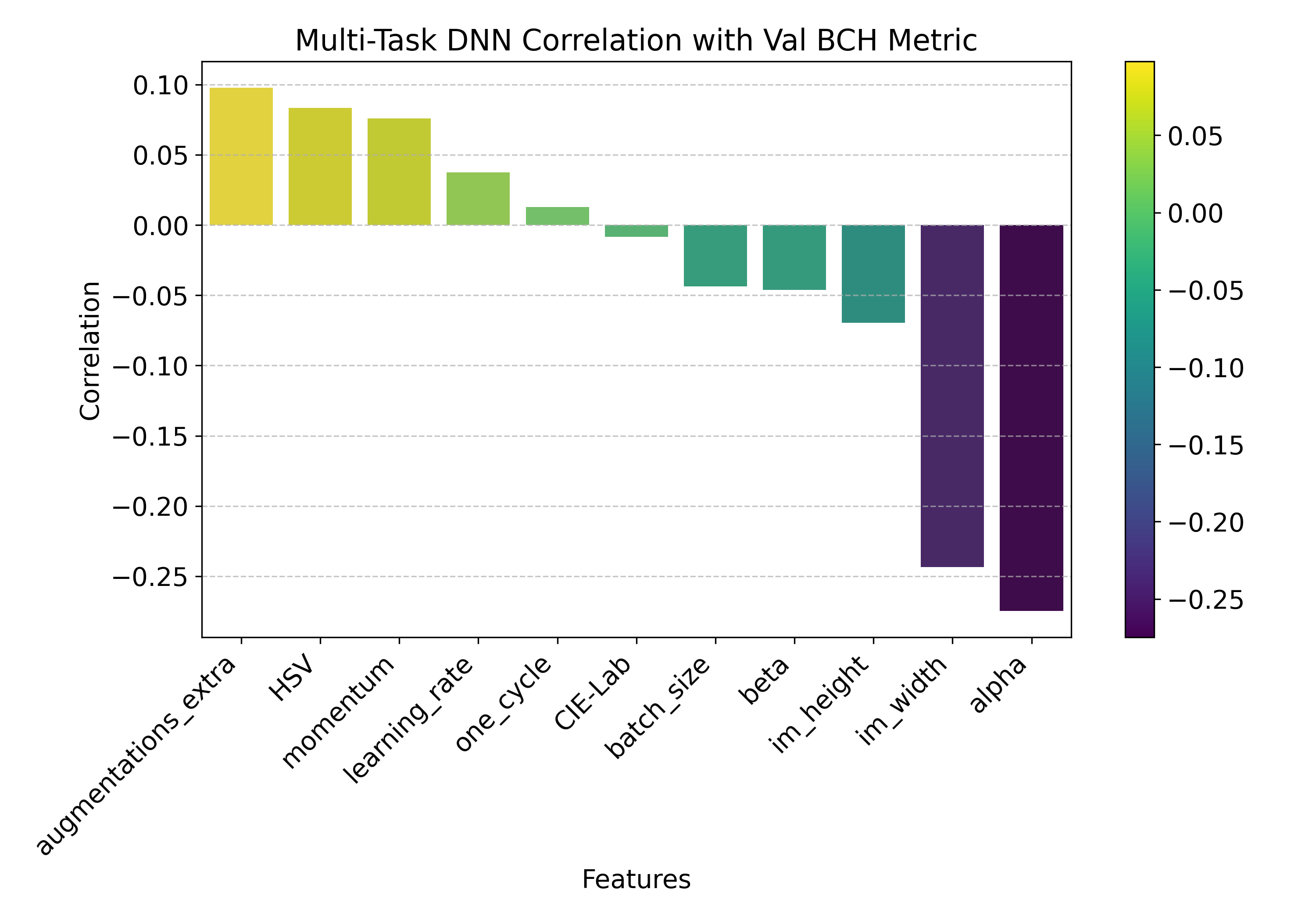}
    \caption{This plot shows an example of the correlation coefficient with the Validation BCH metric of the best $25\%$ of the runs trained on the \Ssj dataset split.}
    \label{fig:multi_correlation_S2S}
\end{figure}

Given the huge parameter space, this time the sweep was performed with random sampling, with a total run count greater than 4000. 
To understand which parameters were most influential, we analyzed the top 25\% of all experimental runs.  An example of the correlation between some of the hyper-parameters and the BCH validation score for this subset is visualized in Fig.~\ref{fig:multi_correlation_S2S}. The full analysis revealed that standard deep learning parameters like learning rate and regularization weights, along with image dimensions, were the most influential factors on performance. The detailed hyper-parameter configurations for the best models are available in our GitHub repository and the complete list of experimental run can be found on the Weights\&Biases report.\footnote{
\url{https://api.wandb.ai/links/sparviero19/s2p10p3a}}

\section{Multi-Task DNN Hyper-parameter Validation Metric} \label{app:hyper-params-validation}

Hyperparameter selection in multi-task learning requires identifying a single metric $\mathcal{S}$ that effectively summarizes performance across different models, which is particularly challenging when tasks differ significantly.

For SSC estimation, mean absolute error (MAE) on the °Brix scale provides an intuitive measure. Color estimation, however, is better assessed using Balanced Accuracy ($A_{color}$) due to the categorical nature of the target variable (although the cross-entropy loss is also applicable it is less interpretable). We combine these two metrics to evaluate the overall performance of the estimators. However, since the metrics are in different scales and units we normalize them using a factor \( \eta \).

Our evaluation criteria also align with agronomic quality standards, emphasizing the importance of avoiding the harvest of unripe bunches, while allowing for some variation in quality among ripe bunches. This is achieved by introducing hinge-like penalties that apply only when estimations fall below predefined quality thresholds for SSC ($t_{brix}$) and color ($t_{color}$) estimation, ensuring that models are primarily evaluated on their ability to meet minimum quality standards.
The final score, \( \mathcal{S} \) integrates the normalized performance metrics and the agronomic quality criteria. We call this metric Brix-Color-Hinge (BCH) metric.

The formal definition of the score $\mathcal{S}$ is the following:
\begin{align}
    \mathcal{S} &= \eta (1-A_{color}) + \mathcal{L}_{brix} +\mathcal{L}_{CH} + \mathcal{L}_{BH} \\
    L_{BH} &= \max(0, L_{brix} - t_{brix}))^\rho \\
    L_{CH} &= \max(0, - \eta(A_{color} - t_{color}))^\rho 
\end{align}    

where $\rho$ is the degree of the two hinge penalties $\mathcal{L}_{BH}$ (for °Brix) and $\mathcal{L}_{CH}$ (for color), set to 2 in all our experiments. $t_{brix}$ and $t_{color}$ represent the agronomic quality thresholds for °Brix and color, respectively, and $\eta$ is the normalization factor between SSC and color scores. In our experiments, $t_{brix}=2.5$ °Brix, which is comparable with the standard deviation of the training data, and $t_{color}=0.8$, since an error rate of $80\%$ is usually considered acceptable for human performance in this context \citet{canopies_D2_4}.
The normalization factor is empirically set as
\begin{equation}
    \eta = \frac{t_{brix}}{t_{color}} .
\end{equation}

\end{document}